\documentclass{SCIS2021}
\RequirePackage{xspace}

\makeatletter
\DeclareRobustCommand\onedot{\futurelet\@let@token\@onedot}
\def\@onedot{\ifx\@let@token.\else.\null\fi\xspace}

\def\eg{\emph{e.g}\onedot} 
\def\ie{\emph{i.e}\onedot} 
 
\def\etc{\emph{etc}\onedot} 
\def\wrt{w.r.t\onedot} 
 
\def\etal{\emph{et al}\onedot}
\makeatother

\definecolor{Blue}{RGB}{0,176,240}
\definecolor{Green}{RGB}{0,176,80}

\usepackage[capitalize]{cleveref}
\crefname{section}{Sec.}{Secs.}
\Crefname{section}{Section}{Sections}
\Crefname{table}{Table}{Tables}
\crefname{table}{Tab.}{Tabs.}
\Crefname{equation}{Equation}{Equations}
\crefname{equation}{Eqn.}{Eqns.}

\DeclareMathOperator*{\argmin}{argmin}

\usepackage{lipsum}
\usepackage{bm}
\usepackage{overpic}
\usepackage{caption}
\usepackage{tabularx}

\usepackage[table]{xcolor}

\begin{document}
\ArticleType{REVIEW}
\Year{2020}
\Month{}
\Vol{}
\No{}
\DOI{}
\ArtNo{}
\ReceiveDate{}
\ReviseDate{}
\AcceptDate{}
\OnlineDate{}

\title{A Survey on Leveraging Pre-trained\\Generative Adversarial Networks\\for Image Editing and Restoration}{A Survey on Leveraging Pre-trained Generative Adversarial Networks for Image Editing and Restoration}

\author[1,2]{Ming LIU}{}
\author[1]{Yuxiang WEI}{}
\author[1]{Xiaohe WU}{}
\author[1]{Wangmeng ZUO}{{wmzuo@hit.edu.cn}}
\author[2]{Lei ZHANG}{}

\AuthorMark{Ming LIU}

\AuthorCitation{Ming LIU, Yuxiang WEI, Xiaohe WU, Wangmeng ZUO and Lei ZHANG}


\address[1]{School of Computer Science and Technology, Harbin Institute of Technology, Harbin {\rm 150001}, China}
\address[2]{Department of Computing, Hong Kong Polytechnic University, Hong Kong {\rm 999077}, China}

\abstract{
    Generative adversarial networks (GANs) have drawn enormous attention due to the simple yet effective training mechanism and superior image generation quality.
    With the ability to generate photo-realistic high-resolution (\eg, $1024\times1024$) images, recent GAN models have greatly narrowed the gaps between the generated images and the real ones.
    Therefore, many recent works show emerging interest to take advantage of pre-trained GAN models by exploiting the well-disentangled latent space and the learned GAN priors.
    In this paper, we briefly review recent progress on leveraging pre-trained large-scale GAN models from three aspects, \ie,
    1)~the training of large-scale generative adversarial networks,
    2)~exploring and understanding the pre-trained GAN models, and
    3)~leveraging these models for subsequent tasks like image restoration and editing.
    More information about relevant methods and repositories can be found at \url{https://github.com/csmliu/pretrained-GANs}.
}

\keywords{Survey, Generative Adversarial Networks, Pre-trained Models, Image Editing, Image Restoration}

\maketitle

\section{Introduction}
\label{sec:intro}
Generative adversarial networks (GANs)~\cite{GAN} are a representative type of generative models, which push the fake data distribution towards the real one via adversarial learning, making the sampled data (\eg, images) more similar to the real one.
In general, the GAN models are comprised of two parts, \ie, a generator $\mathit{G}$ and a discriminator $\mathit{D}$.
Specifically, the generator aims at generating fake samples as realistic as possible, while the discriminator, in turn, aims at distinguishing between real and fake samples to avoid being fooled by the generator.
During the adversary process, the ability of $\mathit{G}$ and $\mathit{D}$ can be constantly improved.
Finally, they may reach a balanced state (\ie, Nash equilibrium), where $\mathit{G}$ is capable of generating images that $\mathit{D}$ cannot tell between real and fake.
The past few years have witnessed the rapid development of GANs~\cite{GAN}, and a series of works make tremendous efforts to improve the quality of GAN-generated images in terms of network structures~\cite{LapGAN,DCGAN,StackGAN,SAGAN}, loss functions~\cite{LSGAN,BEGAN,RaGAN,WGAN,WGAN-GP,SNGAN}, and so on.
Recently, several GAN models (\eg, PGGAN~\cite{PGGAN}, BigGAN~\cite{BigGAN}, and StyleGAN series~\cite{StyleGAN,StyleGAN2,StyleGAN2-Ada,StyleGAN3}) have shown superior image generation ability to produce photo-realistic high-resolution (\eg, $1024\times1024$) images.

Therefore, the exploration on leveraging GANs for enhancing image quality has drawn upsurging attention.
Prior works often simply take the adversarial loss as an extra term in their learning objectives.
On the one hand, the discriminator and adversarial loss can serve as out-of-the-box components, which could be readily incorporated into many other models, and bring no extra complexity during inference.
On the other hand, applying adversarial training generally can avoid the models being optimized to a blurry average solution, which makes the output image contain more sharp and clear details.
As such, the effectiveness has been verified in many vision tasks such as image editing~\cite{Pix2pix,CycleGAN,StarGAN,AttGAN,STGAN,StarGANv2}, image super-resolution~\cite{SRGAN,ESRGAN,Real-ESRGAN,BSRGAN}, and image deblurring~\cite{DeblurGAN,DeblurGANv2,DBGAN}.

However, with the development of GANs, it becomes much harder for such a utilization scheme to fully explore the potential of GAN models.
Therefore, many recent works turn to explore and leverage pre-trained GAN models, whose advantages are analyzed as follows.

\begin{figure}[!t]
    \centering
    \footnotesize
    \begin{tabular}{cccccc}
        \includegraphics[width=0.16\textwidth]{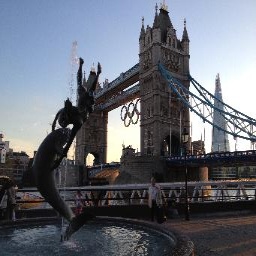} \hspace{-2.75mm} & \hspace{-2.75mm}
        \includegraphics[width=0.16\textwidth]{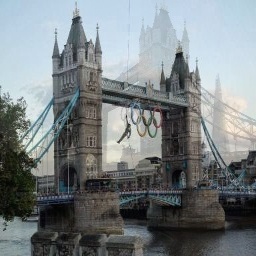} \hspace{-2.75mm} & \hspace{-2.75mm}
        \includegraphics[width=0.16\textwidth]{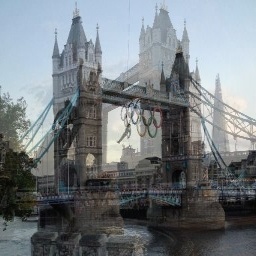} \hspace{-2.75mm} & \hspace{-2.75mm}
        \includegraphics[width=0.16\textwidth]{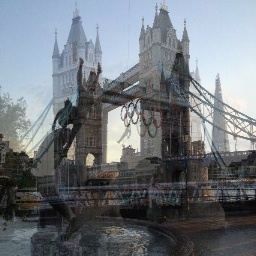} \hspace{-2.75mm} & \hspace{-2.75mm}
        \includegraphics[width=0.16\textwidth]{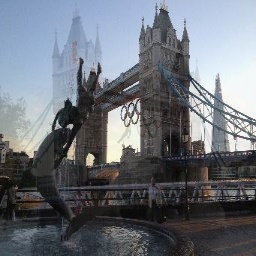} \hspace{-2.75mm} & \hspace{-2.75mm}
        \includegraphics[width=0.16\textwidth]{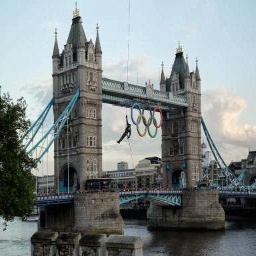} \\[-2mm]
        Image $\textbf{x}_1$ \hspace{-2.75mm} & $\alpha=0.2$ \hspace{-2.75mm} & $\alpha=0.4$ \hspace{-2.75mm} & $\alpha=0.6$ \hspace{-2.75mm} & $\alpha=0.8$ \hspace{-2.75mm} & Image $\textbf{x}_2$ \\
        \includegraphics[width=0.16\textwidth]{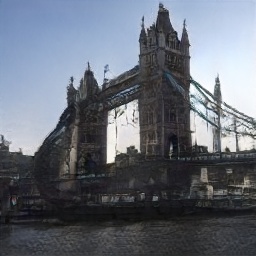} \hspace{-2.75mm} & \hspace{-2.75mm}
        \includegraphics[width=0.16\textwidth]{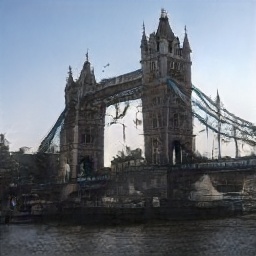} \hspace{-2.75mm} & \hspace{-2.75mm}
        \includegraphics[width=0.16\textwidth]{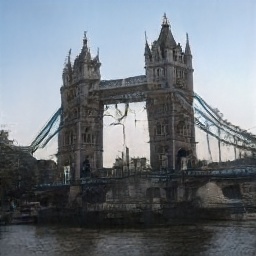} \hspace{-2.75mm} & \hspace{-2.75mm}
        \includegraphics[width=0.16\textwidth]{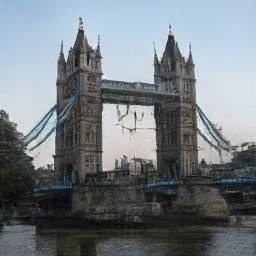} \hspace{-2.75mm} & \hspace{-2.75mm}
        \includegraphics[width=0.16\textwidth]{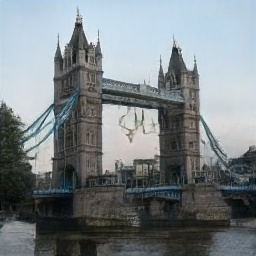} \hspace{-2.75mm} & \hspace{-2.75mm}
        \includegraphics[width=0.16\textwidth]{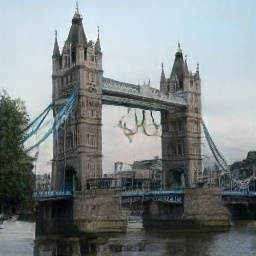} \\[-2mm]
        $G(E(\textbf{x}_1))$ \hspace{-2.75mm} & $\beta=0.2$ \hspace{-2.75mm} & $\beta=0.4$ \hspace{-2.75mm} & $\beta=0.6$ \hspace{-2.75mm} & $\beta=0.8$ \hspace{-2.75mm} & $G(E(\textbf{x}_2))$
    \end{tabular}
    \caption{Image interpolation in the pixel space and the latent space. $\textbf{x}_1$ and $\textbf{x}_2$ are two images of the same object, while $G(E(\mathbf{x}))$ means the inversion result of $\mathbf{x}$. The first row shows the results of interpolating directly in the pixel space via $\mathbf{x}=\alpha\cdot\mathbf{x}_1+(1-\alpha)\cdot\mathbf{x}_2$. In contrast, the second row shows the results of interpolating in the latent space via $\mathbf{x}=G(\beta\cdot\mathit{E}(\mathbf{x}_1)+(1-\beta)\cdot\mathit{E}(\mathbf{x}_2))$. It can be seen that the interpolation in the latent space is semantically more plausible. The sample images are from the tower subset of LSUN dataset~\cite{LSUN}, and the results in the second row are generated via the officially released model of IDInvert~\cite{IDInvert}.}
    \label{fig:intro_interpolation}
\end{figure}

\vspace{0.5em}
\noindent\textbf{Disentanglement of GAN Latent Space.}
For many vision tasks like image editing, it is a basic problem to identify image contents and the manipulation mode of a concept to be edited.
Since the concepts are entangled with each other and form an interlaced group of editing spaces (\eg, changing a facial image to \textit{smile} or \textit{old} can both cause wrinkles).
Even with the well-labeled datasets (\eg, CelebA~\cite{CelebA}), it is still a very challenging task to describe the manipulation mode directly on the pixels~\cite{StarGAN,AttGAN,STGAN,StarGANv2}.
As a remedy, pre-trained GAN models provide yet another possible solution, where the disentanglement of latent spaces has been explored recently~\cite{StyleGAN,voynov2020unsupervised}.
As shown in \cref{fig:intro_interpolation}, interpolation between images will cause obvious artifacts, while the same operation in the latent space will be semantically more plausible.
Such a phenomenon clearly illustrates the latent space disentanglement ability of recent GAN models.

\vspace{0.5em}
\noindent\textbf{Leveraging pre-trained GAN Priors.}
For tasks with image outputs, regularization and natural image priors are often applied to enhance the output image quality, \eg, local smoothness~\cite{rudin1992nonlinear}, non-local self-similarity~\cite{buades2005non}, sparsity~\cite{elad2006image}, \etc.
Nevertheless, they are either hand-crafted with manually tuned parameters, or based on strong assumptions that are usually not applicable in real applications, making the performance and flexibility greatly limited.
On the contrary, the learning objective of GAN models is to generate high-quality diverse images from the latent code (which is generally a random noise).
In other words, given a vector located in the latent space, the well-trained GAN models are very likely to generate a high-quality output.
Therefore, the pre-trained GAN models are naturally able to provide stronger and more reliable high-quality natural image priors.
Furthermore, the GAN priors can be helpful to a wide range of tasks, making it with much more practical significance.

\vspace{0.5em}
\noindent\textbf{Training Scheme \& Development Potentials.}
Collecting pairwise training data is extremely time-consuming and expensive, and it is impossible to collect abundant training data for all tasks.
Fortunately, GAN models are trained in an unsupervised manner, where the training images are much easier to collect without requiring ground-truth labels or reference images.
Indeed, there are already a huge amount of unlabeled images, and the unsupervised training scheme of GAN models makes it easier to apply the extremely large-scale training sets and to train models with greater capacity for better performance.

As a result, many recent works show rising interest on pre-trained GAN models and have developed a series of methods.
In this paper, we review the recent progress on leveraging pre-trained GAN models, from understanding GAN models~\cite{GANDissection,GANSeeing,GANalyze,GANSpace} to leveraging them for subsequent tasks such as image editing~\cite{suzuki2018spatially,GANPaint,StyleRig,StyleFlow} and restoration~\cite{PULSE,pSp,GLEAN,GFPGAN,GPEN}, mainly involving,
\begin{itemize}
    \item We survey the characteristics of pre-trained GAN models, and the efforts for exploring and acquiring these characteristics are also introduced.
    \item Focusing on image restoration and editing tasks, we summarize and compare relevant methods to utilize pre-trained GAN models.
    \item We discuss the open problems and potential research directions in this field.
\end{itemize}

The rest of this paper is organized as follows.
To begin with, \cref{sec:GANs} introduces recent progress of GAN models, including network architecture, training mechanism, and so on.
Then, by exploring the characteristics of pre-trained GAN models (\eg, latent space exploration), the relevant methods and experiments for understanding pre-trained GAN models are reviewed in \cref{sec:understanding}.
Based on the impressions of latent space, we introduce and summarize the methods in \cref{sec:applications}, which leverage pre-trained GANs for subsequent applications, especially the image editing and restoration tasks.
Finally, we discuss the open problems in this field, and point out some potential directions for future research in \cref{sec:discussion}.

\section{Pre-trained Generative Adversarial Networks}
\label{sec:GANs}

\subsection{Preliminary}
\label{sec:GANs_preliminary}

In the deep learning~\cite{deep_learning} era, when researchers are referring to a task, generally there accompany two fundamental problems, \ie, \textit{data sets} and \textit{evaluation metrics}.
For GAN models, these two problems are even more important.
On the one hand, numerous training samples are typically required to avoid over-fitting and to tap the potential of their superior capacity.
On the other hand, the evaluation of generative models is more difficult since there are no ground-truth or labels for assessment, and generating thousands or even millions of images for user study is a mission impossible due to the time and labor costs.
Therefore, before introducing the large-scale GAN models, we brief the relevant datasets, as well as the metrics for evaluating the pre-trained GAN models.
Besides, we summarize the symbols used in this survey in \cref{tab:symbols}.

\begin{table}[t]
    \centering
    \caption{A summary on the symbols used in this survey.}\label{tab:symbols}
    \scalebox{0.54}{
    \begin{tabular}{ccc}
        \toprule
        Format & Definition & Examples \\
        \midrule
        Upper-case italic letters (\eg, $\mathit{G}$) & Models or networks & Generator $\mathit{G}$; discriminator $\mathit{D}$; encoder $\mathit{E}$; MLP $\mathit{F}$; segmentation model $\mathit{S}$. \\
        Bold letters (\eg, $\mathbf{z}$) & Images or tensors & Image $\mathbf{x}$, $\mathbf{I}$; latent code (or random noise) $\mathbf{z}$, $\mathbf{w}$, $\mathbf{p}$, $\mathbf{n}$, \etc; learned input $\mathbf{m}$; Fourier feature input $\mathbf{F}$; condition $\mathbf{c}$. \\
        Handwritten letter (\eg, $\mathcal{Z}$) & Latent spaces or distributions     & Latent spaces $\mathcal{Z}$, $\mathcal{C}$, $\mathcal{W}$, $\mathcal{S}$, $\mathcal{P}$, $\mathcal{N}$, $\mathcal{F}$; loss functions $\mathcal{L}$ \\
        Greek letters (\eg, $\alpha$) & Parameters or specified & Interpolation parameters or hyperparameters  $\alpha$, $\beta$, $\lambda$; $\bm{\theta}$ model parameters; any possible network inputs $\bm{\delta}$. \\
        Double struck letter (\eg, $\mathbb{R}$) & Following mathematical definitions & Mathematical expectation $\mathbb{E}$; Spatial dimension $\mathbb{R}$. \\
        \bottomrule
    \end{tabular}}
\end{table}

\subsubsection{Image Datasets for Training Large-scale GAN Models}

\vspace{0.5em}
\noindent\textbf{ImageNet~\cite{ImageNet}} is the very first large-scale natural image dataset, which contains over 14M images from more than 21K classes (referred to as ImageNet-21K), and the average image size is $469\times387$.
The commonly used configuration is a 1,000-class subset with $\sim$1K images per class (denoted by ImageNet-1K), and the images are usually resized to $224\times224$ or $256\times256$ in practice.

\vspace{0.5em}
\noindent\textbf{CelebA~\cite{CelebA}} (Large-scale CelebFaces Attributes Dataset) contains more than 200K facial images from 10k celebrities, each of the images is annotated with 40 attributes, and the aligned images are resized and cropped to $178\times218$.
A subset of CelebA is post-processed to form the CelebA-HQ~\cite{PGGAN} dataset, containing 30K facial images with the resolution of $1024\times1024$.
The dataset is also extended with annotations like CelebAMask-HQ~\cite{MaskGAN} via pixel-wise facial component labeling (face parsing).

\vspace{0.5em}
\noindent\textbf{LSUN~\cite{LSUN} \& AFHQ~\cite{StarGANv2}} are another two datasets of scene and object categories.
The LSUN dataset is composed of 10 scene categories and 20 object categories, each of which contains around 1M labeled images, and the images are provided by resizing the shorter edge to 256 pixels and compressing to JPEG image quality of 75.
On the contrary, AFHQ (Animal Faces-HQ Dataset) is composed of only three categories (\ie, cat, dog, and wildlife), each of which contains around 5K high-quality animal face images with a resolution of $512\times512$.
Recent methods often take a specific category of these two datasets for training their GAN models, \eg, \textit{bedroom}, \textit{church} from LSUN and \textit{cat} from AFHQ.

\vspace{0.5em}
\noindent\textbf{FFHQ~\cite{StyleGAN}} (Flickr-Faces-HQ Dataset) is a high-quality image dataset collected from Flickr\footnote{\url{https://www.flickr.com/}}, containing 70K automatically aligned high-resolution ($1024\times1024$) facial images with diverse data distribution.
To date, FFHQ is the most popular dataset for training GAN models for facial image generation. Image editing and restoration methods based on pre-trained GANs are often trained on FFHQ dataset and evaluated on CelebA-HQ~\cite{PGGAN} to show the generalization ability.

\vspace{0.5em}
\noindent\textbf{Other Datasets}
Apart from the aforementioned ones, some other datasets~\cite{MNIST,SVHN,CIFAR,StyleGAN2-Ada,DeepFashion,Cityscapes} are also employed for prototyping and/or training large-scale GAN models.
To sum up, large scale and favorable image quality are two key ingredients of datasets for training high-quality large-scale GAN models.
Therefore, some other datasets like Objects 365~\cite{Objects365}, Places 365~\cite{Places365}, and Open Images~\cite{OpenImages} may also be used for training GAN models with greater capacity.

\subsubsection{Evaluation Metrics for Assessing GAN models}

\vspace{0.5em}
\noindent\textbf{Mean Opinion Score} (MOS) is the most intuitive index to assess the perceptual quality of images, since the image quality assessment is conducted by human raters.
However, assessing via MOS is very expensive, and the results may be biased due to factors like subjective perceiving differences.
Furthermore, the time-consuming procedure makes it suitable for small-scale assessment, \eg, user study, but hard to be employed for the evaluation during training and broader comparison.

\vspace{0.5em}
\noindent\textbf{Inception Score~\cite{InceptionScore}} (IS) is a commonly used metric for evaluating the image quality and class diversity.
Considering a well-trained classifier (\eg, Inception-v3~\cite{Inception-v3} trained on ImageNet~\cite{ImageNet} for calculating IS), the high confidence of classifying a generated image $\mathbf{\hat{x}}$ into a class $\mathit{c}$ (\ie, smaller entropy of $\mathit{p}(\mathit{c}|\mathbf{\hat{x}})$) generally implies decent image generation quality.
On the other hand, better class diversity requires that the marginal probability distribution of class label approximates uniform distribution, \ie, higher entropy of $\mathit{p}(\mathit{c})$.
Therefore, the Inception Score is defined by
\begin{equation}
    \mathit{IS} = \exp(\mathbb{E}_{\mathbf{\hat{x}}\sim\mathit{p}_\mathit{g}}\mathit{D}_\mathit{KL}(\mathit{p}(\mathit{c}|\mathbf{\hat{x}})\|\mathit{p}(c))),
\end{equation}
where $\mathbf{\hat{x}}$ is sampled from the generated image distribution $\mathit{p}_\mathit{g}$, $\mathit{D}_\mathit{KL}(\cdot\|\cdot)$ denotes Kullback-Leibler (K-L) divergence.
However, IS is unable to detect mode collapse, \eg, a high Inception Score will be derived when the generator produces one and only one high-quality image for each class.

\vspace{0.5em}
\noindent\textbf{Fr\'{e}chet Inception Distance~\cite{FID}}
Considering the drawbacks of IS, Fr\'{e}chet Inception Distance (FID) takes the real images into consideration and calculates the distance between generated images and real ones.
Specifically, FID is defined by
\begin{equation}
    \mathit{FID}=\|\mu_\mathit{r}-\mu_\mathit{g}\|_2^2+\mathit{Tr}(\Sigma_\mathit{r}+\Sigma_\mathit{g}-2(\Sigma_\mathit{r}\Sigma_\mathit{g})^{\frac{1}{2}}),
\end{equation}
where $\mathit{Tr}(\cdot)$ denotes trace of a matrix.
The real and generated image features extracted by Inception-v3~\cite{Inception-v3} are both assumed to follow the multivariate Gaussian distribution, \ie, $\phi(\mathbf{x})\sim\mathcal{N}(\mu_\mathit{r},\Sigma_\mathit{r})$,  $\phi(\mathbf{\hat{x}})\sim\mathcal{N}(\mu_\mathit{g},\Sigma_\mathit{g})$.
By utilizing the real samples, FID is not restricted by the training dataset (\ie, ImageNet~\cite{ImageNet} for Inception-v3~\cite{Inception-v3}), and is much more sensitive to mode collapse.

\vspace{0.5em}
\noindent\textbf{Sliced Wasserstein Distance~\cite{SWD}}
It is worth noting that the Gaussian assumption of FID does not always hold true.
To avoid such assumption, Bonneel \etal~\cite{SWD} proposed to calculate the Sliced Wasserstein distance (SWD) to approximate the Wasserstein distance, \ie,
\begin{equation}
    \mathit{SWD}(\mathbf{I}_\mathit{x},\mathbf{I}_\mathit{y})=(\int_{\mathbb{S}^{\mathit{d}-1}}\mathit{W}^\mathit{p}_\mathit{p}(\mathcal{R}\mathbf{I}_\mathit{x}(.,\theta),\mathcal{R}\mathbf{I}_\mathit{y}(.,\theta))\mathrm{d}\theta)^{\frac{1}{\mathit{p}}},
    \label{eqn:SWD}
\end{equation}
where $\mathbb{S}^{\mathit{d}-1}$ is the unit sphere in $\mathbb{R}^\mathit{d}$, $p$ is set to 2 in practice, $\mathcal{R}$ denotes the Radon transform.
Please refer to \cite{GSWD} for more details.

\vspace{0.5em}
\noindent\textbf{GAN-train \& GAN-test~\cite{GAN-train}}
Another problem in GAN training is over-fitting, \ie, the model memorizes the training samples for generation.
The GAN-train and GAN-test indices provide a more comprehensive evaluation on GAN model.
In particular, GAN-train is the accuracy of a classifier trained on a generated dataset $\mathcal{D}_\mathit{g}$ and evaluated on a real-image validation dataset $\mathcal{D}_\mathit{r}^\mathit{val}$, while GAN-test is the accuracy of a classifier trained on real-image training set $\mathcal{D}_\mathit{r}^\mathit{train}$ and tested on $\mathcal{D}_\mathit{g}$.
These two metrics can identify problems such as mode dropping, poor image quality, over-fitting, \etc.

\vspace{0.5em}
\noindent\textbf{Fr\'{e}chet Segmentation Distance~\cite{GANSeeing}}
As shown in \cite{GANSeeing}, the mode collapse occurs not only at the distribution level, but also at the instance level.
In particular, certain objects may not be generated by a GAN model, and this phenomenon is termed by mode dropping.
For assessing the mode dropping in a GAN model, the Fr\'{e}chet Segmentation Distance (FSD) is defined by modifying FID, \ie,
\begin{equation}
    \mathit{FSD}=\|\mu_\mathit{g}-\mu_\mathit{t}\|_2^2+\mathit{Tr}(\Sigma_\mathit{g}+\Sigma_\mathit{t}-2(\Sigma_\mathit{g}\Sigma_\mathit{t})^{\frac{1}{2}}),
\end{equation}
where $\mu_\mathit{t}$ ($\mu_\mathit{g}$) is the mean pixel count for each object class over a sample of training images (generated images), and $\Sigma_\mathit{t}$ ($\Sigma_\mathit{g}$) is the covariance of these pixels.

Besides, many other metrics like Perceptual Path Length~\cite{StyleGAN}, Linear Separability~\cite{StyleGAN}, Precision and Recall~\cite{PrecRecall}, and Geometry Score~\cite{GeometryScore} are also explored for better evaluating GANs. We recommend \cite{GAN_Metrics} for a more comprehensive survey on GAN metrics.

\begin{figure}
    \centering
    \begin{overpic}[width=.95\linewidth]{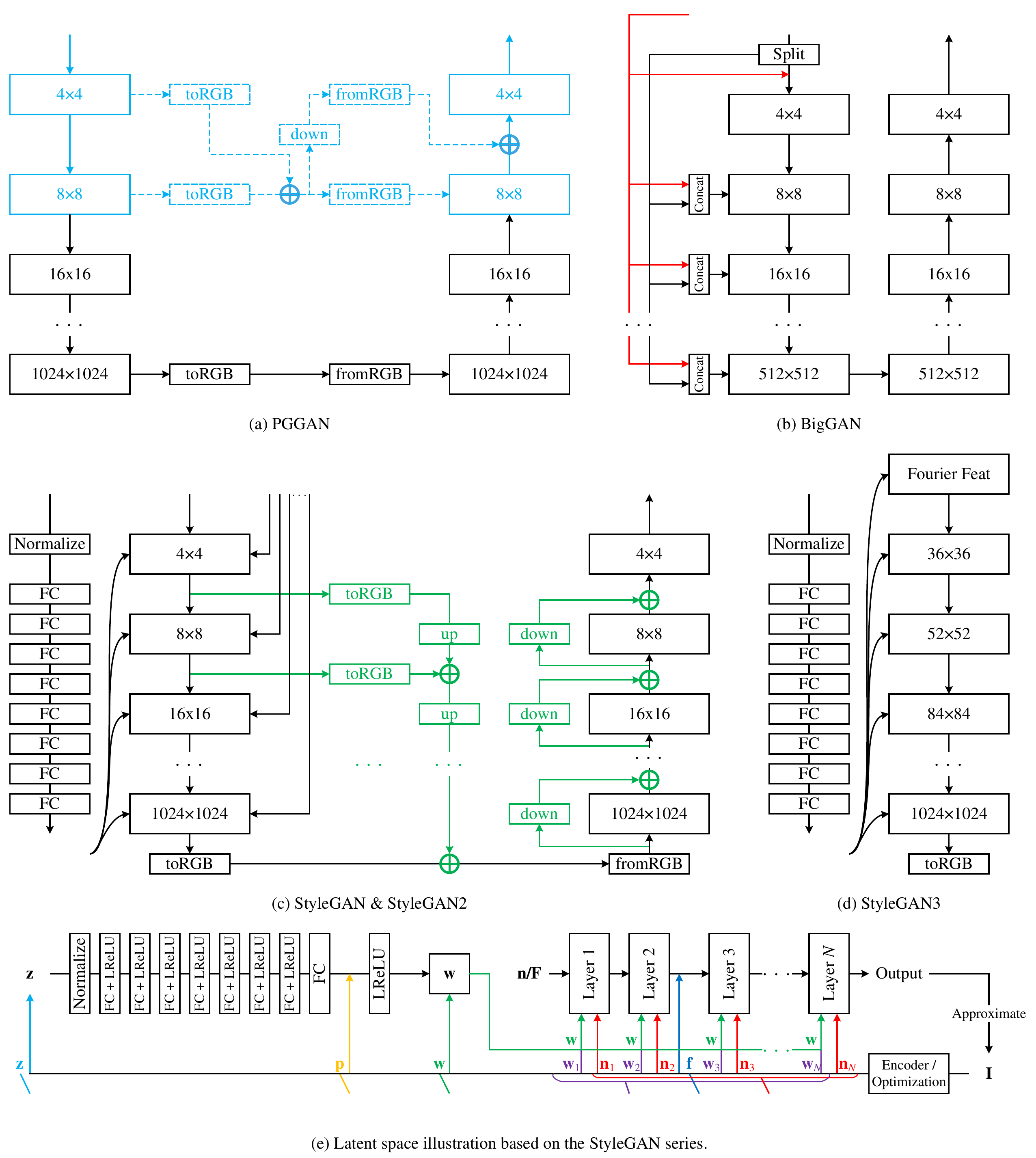}
        \tiny
        \put(3.5, 98.5){\color{Blue}$\mathbf{z}\!\in\!\mathbb{R}^{512}$}
        \put(42.2, 98.5){\color{Blue}Score}
        \put(22.5, 83.5){\color{Blue}$\alpha$}
        \put(21.5, 85.5){\color{Blue}$1\!-\!\alpha$}
        \put(43, 85.5){\color{Blue}$\alpha$}
        \put(38.5, 88){\color{Blue}$1\!-\!\alpha$}

        \put(60, 98.5){\color{red}$\mathbf{c}\!\in\!\mathbb{R}^{128}$}
        \put(66, 98.5){$\mathbf{z}\!\in\!\mathbb{R}^{160}$}
        \put(80, 98.5){Score}

        \put(56.5, 95.5){$\mathbb{R}^{140}$}
        \put(56.5, 93.8){\color{red}$\mathbb{R}^{128}$}
        \put(68.5, 92.5){$\mathbb{R}^{128+20}$}

        \put(56.5, 84.3){\color{red}$\mathbb{R}^{128}$}
        \put(56.8, 82.6){$\mathbb{R}^{20}$}

        \put(56.5, 77.4){\color{red}$\mathbb{R}^{128}$}
        \put(56.8, 75.7){$\mathbb{R}^{20}$}

        \put(56.5, 68.8){\color{red}$\mathbb{R}^{128}$}
        \put(56.8, 67.1){$\mathbb{R}^{20}$}

        \put(2, 58.5){$\mathbf{z}\!\in\!\mathbb{R}^{512}$}
        \put(12.5, 58.5){$\mathbf{m}\!\in\!\mathbb{R}^{4\!\times\!4\!\times\!512}$}
        \put(22.2, 58.5){$\mathbf{n}\!\in\!\mathbb{R}^{\mathit{H}\!\times\!\mathit{W}}$}

        \put(2, 26){$\mathbf{w}\!\in\!\mathbb{R}^{512}$}
        \put(54.3, 58.5){Score}

        \put(67.5, 58.5){$\mathbf{z}\!\in\!\mathbb{R}^{512}$}
        \put(67.5, 26){$\mathbf{w}\!\in\!\mathbb{R}^{512}$}

        \put(0, 4){$\mathcal{Z}$ Space}
        \put(27.5, 4){$\mathcal{P}$ Space}
        \put(36, 4){$\mathcal{W}$ Space}
        \put(50.5, 4){$\mathcal{W}+$ Space}
        \put(57.9, 4){$\mathcal{F}$ Space}
        \put(63.7, 4){$\mathcal{N}$ Space}

    \end{overpic}
    \caption{Illustration of recent GAN models (see (a)$\sim$(d)) and the latent spaces of StyleGAN series~\cite{StyleGAN} (see (e)). (a)~For PGGAN~\cite{PGGAN}, the {\color{Blue}blue part} denotes the progressive growing procedure from $4\times4$ to $8\times8$. The components with dash lines are employed for the fade-in strategy, where $\alpha$ is gradually growing to 1. They are discarded when the model grows to a higher-resolution. (b)~For BigGAN~\cite{BigGAN}, a specific noise is delivered to each layer together with the class embedding, and the model is end-to-end trained without the progressive growing procedure. (c)~For StyleGAN~\cite{StyleGAN}, a series of FC layers are deployed to map $\mathbf{z}$ into $\mathbf{w}$. The {\color{Green}green part} only belongs to StyleGAN2. (d)~For StyleGAN3~\cite{StyleGAN3}, the generator is largely modulated to improve the translational and rotation equivariance. The discriminator is omitted since it is identical with that used in StyleGAN2~\cite{StyleGAN2}. (e)~For simplicity, here we take the StyleGAN series as an example to show the latent spaces based on GAN inversion task.}
    \label{fig:GAN_arch}
\end{figure}
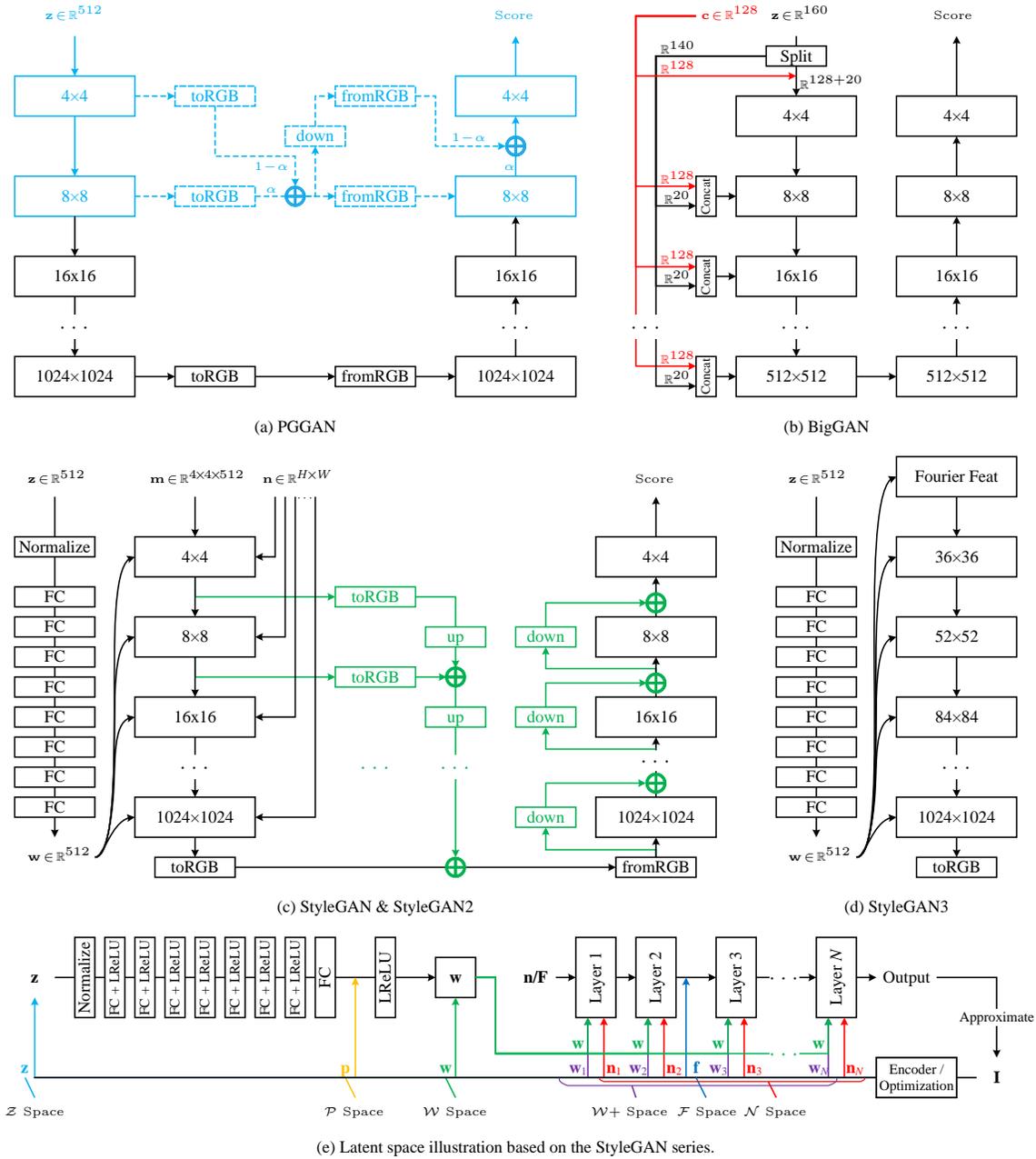

\subsection{Large-scale GAN models}
\label{sec:GANs_development}

GANs are originally proposed for image generation.
As such, we take image generation as an example to introduce the basic concepts and training schemes of GANs, which could be naturally generalized to other domains.
A typical GAN model~\cite{GAN,DCGAN} is composed of a generator and a discriminator, which are denoted by $\mathit{G}$ and $\mathit{D}$, respectively.
Given a random vector $\mathbf{z}\in\mathcal{Z}$, the generator maps it to an image $\mathbf{\hat{x}}$,
\begin{equation}
    \mathbf{\hat{x}}=\mathit{G}(\mathbf{z}, \bm{\delta};\bm{\theta}_\mathit{G}),
    \label{eqn:GAN}
\end{equation}
where $\bm{\delta}$ denotes other possible inputs (\eg, class condition, layer-wise noise, \etc), $\bm{\theta}_\mathit{G}$ represents the parameters of $\mathit{G}$, $\mathbf{\hat{x}}\in\mathbb{R}^{\mathit{H}\times\mathit{W}\times\mathit{C}}$ denotes the generated $\mathit{C}$-channel image with the size of $\mathit{H}\times\mathit{W}$.
Then, the discriminator $\mathit{D}$ takes $\mathbf{\hat{x}}$ (or a real image $\mathbf{x}$) as input, and predicts the probability that the input is from the distribution of real images or the distance to the real image distribution\footnote{Precisely speaking, the metric is \textit{diversity} rather than \textit{distance} for some adversarial loss functions such as vanilla GAN~\cite{GAN}.}, and the learning objective is defined by,
\begin{equation}
    \min_\mathit{G}\max_\mathit{D}\mathcal{L}_\mathit{GAN}(\mathit{G},\mathit{D})=\mathbb{E}_{\mathbf{x}\sim\mathit{p}_\mathit{data}(\mathbf{x})}[\log\mathit{D}(\mathbf{x})]+\mathbb{E}_{\mathbf{z}\sim\mathit{p}_\mathit{\mathbf{z}}(\mathbf{z})}[\log(1-\mathit{D}(\mathit{G}(\mathbf{z})))].
    \label{eqn:loss_GAN}
\end{equation}
In the following, we mainly introduce a handful of recent GAN models that are deployed in the methods reviewed, and we recommend \cite{GANSurvey,StudioGAN} for a comprehensive review of GANs.

\vspace{0.5em}
\noindent\textbf{PGGAN~\cite{PGGAN}}
Leveraging the progressive growing strategy, PGGAN (or ProGAN) achieves the $1024\times1024$ image generation resolution.
Specifically, at the beginning of the training procedure, a $4\times4$ ``image'' is generated by the initial generator $\mathit{G}^{(4)}$, and the discriminator $\mathit{D}^{(4)}$ accordingly takes $4\times4$ input as well.
Once the model converges at a resolution, another layer is deployed at the end of the generator, which generates images with $2\times$ resolution.
Finally we can obtain $\mathit{G}^{(1024)}$, which could generate high-quality $1024\times1024$ images.
The overall architecture is shown in \cref{fig:GAN_arch}(a).

\vspace{0.5em}
\noindent\textbf{BigGAN~\cite{BigGAN}}
As the name implies, BigGAN analyzes the influence of GAN model size on generation quality and proposes an architecture with more parameters and end-to-end trained with large batch-size.
For more stable training, the top singular value of parameters are used to monitor the mode collapse during training, and a novel regularization term on the generator $\mathit{G}$ is designed according to the analysis.
The $\mathit{R}_1$ zero-centered gradient penalty~\cite{GAN_convergence} is also leveraged in the discriminator $\mathit{D}$.
Besides, to achieve class-conditioned generation, BigGAN delivers the class embedding and noise vector in to each residual block of the generator.
Finally, with the large capacity, BigGAN could generate appealing images with the resolution of $512\times512$.
The overall architecture is shown in \cref{fig:GAN_arch}(b).

\vspace{0.5em}
\noindent\textbf{StyleGAN~\cite{StyleGAN}}
The StyleGAN series~\cite{StyleGAN,StyleGAN2,StyleGAN3} are the most popular GAN models in recent years.
Instead of taking noise as input, Karras~\etal proposed to learn a constant input $\mathbf{m}$ (as shown in \cref{fig:GAN_arch}(c)).
And they used a mapping network composed of several fully-connected layers to map the noise $\mathbf{z}$ into a latent representation $\mathbf{w}$, which is delivered to the AdaIN~\cite{AdaIN} layer in each scale.
In addition, layer-wise per-pixel noise was introduced in each scale for further performance improvement and achieved stochastic variation on generated details.
StyleGAN follows the progressive growing training scheme of PGGAN~\cite{PGGAN}, the final-stage structure of StyleGAN is shown as the black part in \cref{fig:GAN_arch}(c).

\vspace{0.5em}
\noindent\textbf{StyleGAN2~\cite{StyleGAN2}}
The overall structure of StyleGAN2 is similar to StyleGAN, and the block structure and regularization terms are modulated for better generalization quality.
Besides, StyleGAN2 deploys a ``toRGB'' module in each scale for substituting the progressive growing strategy.
The lazy regularization and perceptual path length regularization are also deployed.
Furthermore, Karras~\etal~\cite{StyleGAN2-Ada} proposed StyleGAN2-Ada based on a delicately designed discriminator augmentation mechanism, which largely improves the generated image quality and significantly stabilizes training in limited data regimes.

\vspace{0.5em}
\noindent\textbf{StyleGAN3~\cite{StyleGAN3}}
Since the StyleGAN architecture was proposed, utilizing pre-trained GAN models for downstream tasks has drawn much attention, making the equivariance a necessity for many applications.
However, the synthesis process of previous GAN models depends on the coordinates of pixels rather than the surface of the generated objects.
Karras~\etal analyzed the cause of such conditions, and designed a model with translational equivariance and rotation equivariance, which achieves comparable performance against StyleGAN2 but with even less parameters.

The comparison between these GAN models are shown in \cref{fig:GAN_arch} and \cref{tab:GAN_comparison}.
It can be seen that, in order to achieve realistic generation, these methods keep exploring smoother and more stable \textit{information flow pipelines and architectures}, more disentangled and controllable \textit{input and condition forms}, as well as better \textit{training schemes}.
In the next section, we will delve more deeply into the GAN models, trying to analyze and understand the characteristics and properties of pre-trained GANs.

\begin{table}
    \centering
    \caption{Comparison of recent GAN models. The two numbers of parameters for StyleGAN3~\cite{StyleGAN3} are respectively for the translational equivariant configuration and the rotation equivariant configuration. $\mathbf{z}$, $\mathbf{m}$, and $\mathbf{F}$ denote random noise, learned constant, and Fourier feature input~\cite{FourierFeature}, respectively.}
    \label{tab:GAN_comparison}
    \scalebox{.75}{
        \begin{tabular}{cccccccc}
            \toprule
            GAN Model                  & Publication & Resolution & \# Params   & Training Scheme        & Input        & Condition               & Regularization \\
            \midrule
            PGGAN~\cite{PGGAN}         & ICLR 2018   & 1024       & 23.1M       & Growing                & $\mathbf{z}$ & -                       &  Grad. Penalty (GP)              \\
            BigGAN~\cite{BigGAN}       & ICLR 2019   & 512        & 82.5M       & End-to-end             & $\mathbf{z}$ & $\mathbf{c}+\mathbf{z}$ &  Ortho. Reg. \& $\mathit{R}_1$             \\
            StyleGAN~\cite{StyleGAN}   & CVPR 2019   & 1024       & 26.2M       & Growing                & $\mathbf{m}$ & $\mathbf{w}+\mathbf{n}$ &  Mixing Reg., GP \& $\mathit{R}_1$              \\
            StyleGAN2~\cite{StyleGAN2} & CVPR 2020   & 1024       & 30.0M       & Multi-scale            & $\mathbf{m}$ & $\mathbf{w}+\mathbf{n}$ &  Mixing Reg., Lazy $\mathit{R}_1$ \& PPL              \\
            StyleGAN3~\cite{StyleGAN3} & NeurIPS 2021   & 1024       & 22.3M/15.8M & End-to-end             & $\mathbf{F}$ & $\mathbf{w}$            &  $\mathit{R}_1$             \\
            \bottomrule
    \end{tabular}}
\end{table}

\section{Analyzing and Understanding Large-scale GAN Models}
\label{sec:understanding}
In this section, we first analyze the GAN models from the perspective of image generation, and review the recent progress on understanding the neurons of pre-trained GAN models.
Then, to leverage the priority of the recent GAN architectures (\eg, StyleGAN series~\cite{StyleGAN,StyleGAN2,StyleGAN3}), the frequently used latent spaces are introduced, together with a series of methods for mapping images back to the corresponding latent space (\ie, GAN inversion).
Finally, we explore the current progress on latent space disentanglement, one of the key problems of image manipulation using pre-trained GANs.

\subsection{Neuron Understanding}
\label{subsec:Understanding_neuron}

\subsubsection{Preliminary}
Analogous to other neural networks, GAN~\cite{GAN} models are also trained in a data-driven manner, making them a black box which is hard to understand and interpret.
To open the black box and better understand the image generation process in GAN generators, a series of works~\cite{GANDissection,GANInverting,GANSeeing,GANCorrection} propose to disassemble the pre-trained GAN models and see how objects in the output images are generated.

In particular, as introduced in \cref{sec:GANs_development}, the generation process in the generator $\mathit{G}$ could be written as $\mathbf{\hat{x}}=\mathit{G}(\mathbf{z})$.
Following the main model paradigm in the deep learning era, a typical GAN model is composed of a stack of neural layers (\eg, a $\mathit{N}$-layer GAN model), thus we can decompose the generator $\mathit{G}$ into two parts at the $\mathit{i}$-th layer, and \cref{eqn:GAN} can be rewritten as,
\begin{equation}
    \mathbf{\hat{x}}=\mathit{G}(\mathbf{z})=(\mathit{G}_{\mathit{i}+1}^\mathit{N}\circ\mathit{G}_{1}^\mathit{i})(\mathbf{z})=\mathit{G}_{\mathit{i}+1}^\mathit{N}(\mathit{G}_{1}^\mathit{i}(\mathbf{z})),
\end{equation}
where $\mathit{G}_\mathit{i}$ is the $\mathit{i}$-th layer of $\mathit{G}$, and $\mathit{G}_\mathit{a}^\mathit{b}=\mathit{G}_\mathit{b}\circ\mathit{G}_{\mathit{b}-1}\circ\cdots\circ\mathit{G}_{\mathit{a}+1}\circ\mathit{G}_\mathit{a}$.
Then the features map $\mathbf{f}_\mathit{i}$ at the $\mathit{i}$-th layer (\ie, $\mathbf{f}_\mathit{i}=\mathit{G}_{1}^\mathit{i}(\mathbf{z})$) contains all information for generating the objects in the output images.

\subsubsection{GAN Dissection}
\label{subsubsec:GANDissection}
For a certain convolution layer, the activation at each position is obtained by the sum-of-product operation between the previous features and the convolution kernels.
Since each output channel has a particular kernel, we can see it as a neuron, which potentially controls the generation of some objects, and the discussions in this subsection are based on this observation.

Bau~\etal~\cite{GANDissection} first built the relationship between the generated objects and the feature maps from $\mathbf{f}_\mathit{i}$ via network dissection techniques~\cite{NetworkDissection}.
Specifically, for each object class $\mathit{c}$, a semantic segmentation model $\mathit{S}_\mathit{c}$ is deployed to get the binary segmentation result $\mathit{S}_\mathit{c}(\mathbf{\hat{x}})$.
Then, the relationship between this class and a feature map $\mathbf{f}_\mathit{i,u}$ (aka, a neuron) is evaluated by the \textit{spatial agreement} between the thresholded feature map and the segmentation result, which is evaluated by the intersection-over-union (IoU) measure, \ie,
\begin{equation}
    \mathit{IoU}_\mathit{u,c}=\frac{\mathbb{E}_\mathbf{z}\left|(\mathbf{f}^{\uparrow}_\mathit{i,u} > \mathit{t}_\mathit{u,c})\land\mathit{S}_\mathit{c}(\mathbf{\hat{x}})\right|}{\mathbb{E}_\mathbf{z}\left|(\mathbf{f}^{\uparrow}_\mathit{i,u} > \mathit{t}_\mathit{u,c})\lor\mathit{S}_\mathit{c}(\mathbf{\hat{x}})\right|},\ \mathrm{where}\ \mathit{t}_\mathit{u,c}={\arg\max}_\mathit{t}\frac{\mathbf{I}(\mathbf{f}^{\uparrow}_\mathit{i,u} > \mathit{t};\mathit{S}_\mathit{c}(\mathbf{\hat{x}}))}{\mathbf{H}(\mathbf{f}^{\uparrow}_\mathit{i,u} > \mathit{t};\mathit{S}_\mathit{c}(\mathbf{\hat{x}}))},
\end{equation}
where $\land$ and $\lor$ represent intersection and union operations, $\uparrow$ denotes the upsampling operation, and $\mathit{t}_\mathit{u,c}$ is determined by maximizing the portion of the mutual information $\mathbf{I}$ in the joint entropy $\mathbf{H}$.

\begin{figure}[t]
    \centering
    \footnotesize
    \begin{minipage}{0.475\linewidth}
        \centering
        \begin{tabular}{cccc}
                \hspace{-2.75mm}
                \includegraphics[width=0.24\textwidth]{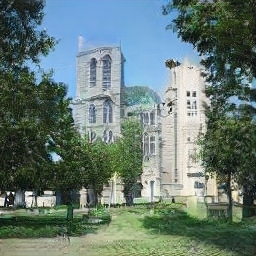} \hspace{-2.75mm} & \hspace{-2.75mm}
                \includegraphics[width=0.24\textwidth]{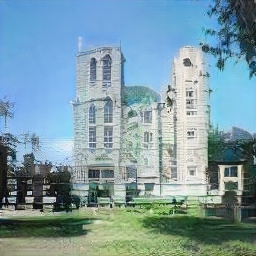} \hspace{-2.75mm} & \hspace{-2.75mm}
                \includegraphics[width=0.24\textwidth]{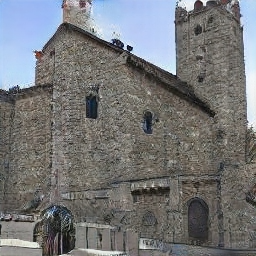} \hspace{-2.75mm} & \hspace{-2.75mm}
                \includegraphics[width=0.24\textwidth]{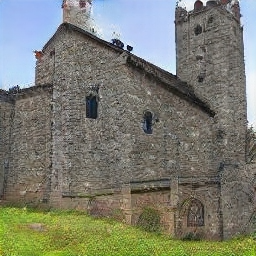} \\[-1.5mm]
                \hspace{-2.75mm}
                Input \hspace{-2.75mm} & \hspace{-2.75mm}
                Remove Trees \hspace{-2.75mm} & \hspace{-2.75mm}
                Input \hspace{-2.75mm} & \hspace{-2.75mm}
                Add Grasses
        \end{tabular}
        \captionof{figure}{Removing or adding particular classes via relevant unit discovery in GANDissection~\cite{GANDissection}.}
        \label{fig:GANDissection}
    \end{minipage}
    \hspace{2mm}
    \begin{minipage}{0.475\linewidth}
        \centering
        \begin{tabular}{cccc}
            \hspace{-2.75mm}
            \includegraphics[width=0.24\textwidth]{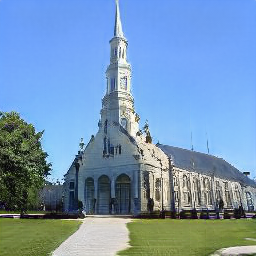} \hspace{-2.75mm} & \hspace{-2.75mm}
            \includegraphics[width=0.24\textwidth]{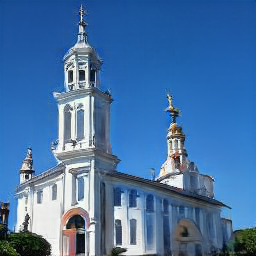} \hspace{-2.75mm} & \hspace{-2.75mm}            \includegraphics[width=0.24\textwidth]{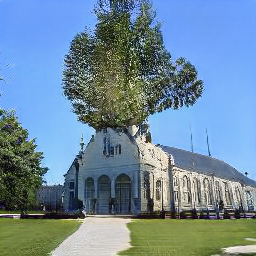} \hspace{-2.75mm} & \hspace{-2.75mm}
            \includegraphics[width=0.24\textwidth]{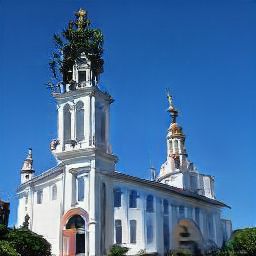} \\[-1.5mm]
            \multicolumn{2}{c}{\hspace{-2.75mm}Two inputs\hspace{-2.75mm}} & \multicolumn{2}{c}{\hspace{-2.75mm} \texttt{church}$\to$\texttt{dome}$\Rightarrow$\texttt{church}$\to$\texttt{tree}}
        \end{tabular}
        \captionof{figure}{Manipulating the geneartion rules via \cite{GANRewriting}, the two input images are modified via the same model.}
        \label{fig:GANRewritting}
    \end{minipage}
\end{figure}

In order to further verify the causal effects between the neurons and the generation results, after identifying the units which are relevant to a specific class of objects, Bau~\etal~\cite{GANDissection} further ablated (or inserted) the units by setting the activation of the relevant neurons to zero (or a class-specific constant).
\cref{fig:GANDissection} shows the results of such modifications.
It can be seen that the whole church appears after ablating the trees, and the grasses are added to the other image.
The result indicates that
1)~the generation is under the control of the relevant neurons, whose activation values determine the absence of the objects in the final generated image, and
2)~the previously unseen objects may still be generated, but sheltered by other objects.
Tousi~\etal~\cite{GANCorrection} further explored this phenomenon in multiple layers, and achieved a more accurate manipulation.
This procedure could be regarded as an inverse procedure of the classification task to some extent, where the final prediction also shelters other classes, and we recommend \cite{ActivationAtlas} for an intuitive and interactive understanding.

\subsubsection{Generation Rule Manipulation}
\label{subsubsec:GANrewriting}
When adding some objects in an image as stated in \cref{subsubsec:GANDissection}, it will fail to generate the desired objects somewhere (\eg, adding a door in the sky).
Since the objects could be successfully inserted in other regions (\eg, adding a door on the wall), Bau~\etal~\cite{Rewriting} asserted that there are some rules in the generation process learned from the data, and they seek to manipulate the rules for verification and a general modification method at the model level.

Specifically, as shown in \cref{fig:GANRewritting}, suppose the rule in a generator is like $\mathbf{I}_\mathtt{dome}=\mathit{G}_{\mathtt{church}\rightarrow\mathtt{dome}}(\mathit{G}_{\mathbf{z}\rightarrow\mathtt{church}}(\mathbf{z}))$, where the generator could be regarded as an associative memory that stores a set of key-value pairs $\{(\mathbf{k}_\mathit{i}, \mathbf{v}_\mathit{i})\}$, and is formulated by
\begin{equation}
    \mathbf{v}_\mathit{i}\approx\mathbf{W}\mathbf{k}_\mathit{i},
\end{equation}
we can modulate the parameter $\mathbf{W}$ (aka the associative memory) to reorganize the key-value pairs.
For example, we could change the rule $\mathtt{church}\rightarrow\mathtt{dome}$ to $\mathtt{church}\rightarrow\mathtt{tree}$, and the results are shown in \cref{fig:GANRewritting}.
The results clearly show the generation process, and help to open the black box of GAN models.

\begin{table}
    \newcolumntype{L}[1]{>{\raggedright\arraybackslash}p{#1}}
    \caption{A summary of GAN inversion and methods leveraging pre-trained GANs for image editing and restoration. For the inversion method, ``O'', ``L'', ``T'' represent optimization-based, learning-based, and training-based (or fine-tuning) methods, while ``/'' means no inversion is performed in this method, and the numbers (without square brackets) are the indices of methods used for inversion in this table. Note that the methods are ordered (roughly) according to publicly accessible time (\eg, the appear time on \href{https://arxiv.org}{ArXiv}, \href{https://openreview.net}{openreview.net}, \href{https://openaccess.thecvf.com/}{CVF Open Access}, \etc.). We have also highlighted several works that are frequently used for GAN inversion or image editing/restoration.}
    \label{tab:methods}
    \centering
    \scalebox{0.52}{
    \begin{tabular}{cccccccc}
        \toprule
        \textbf{No.} & \textbf{Method}                                    & \textbf{Publication} & \textbf{Backbone}            & \textbf{\begin{tabular}[c]{@{}c@{}}Latent\\ Space\end{tabular}} & \textbf{\begin{tabular}[c]{@{}c@{}}Inversion\\ Method\end{tabular}} & \textbf{Dataset$^\ast$} & \textbf{Application$^\dagger$}   \\
        \midrule
        1            & BiGAN~\cite{BiGAN}                                 & ICLR 2017            & /                            & $\mathcal{Z}$                                                   & T                                                                   & MN, IN                  & Inv                              \\
        2            & ALI~\cite{ALI}                                     & ICLR 2017            & /                            & $\mathcal{Z}$                                                   & T                                                                   & CF, SV, CA, IN          & Inv, Int                         \\
        3            & Zhu~\etal~\cite{zhu2016generative}                 & ECCV 2016            & DCGAN                        & $\mathcal{Z}$                                                   & L, O                                                                & SH, LS, PL$^\ddagger$   & Inv, Int, AE                     \\
        \rowcolor{green!20}4            & IcGAN~\cite{IcGAN}                                 & NeurIPSw 2016        & cGAN                         & $\mathcal{Z}$, $\mathcal{C}$                                    & L                                                                   & MN, CA                  & Inv, AT, AE                      \\
        5            & Creswell~\etal~\cite{creswell2018inverting}        & T-NNLS 2018          & DCGAN, WGAN-GP               & $\mathcal{Z}$                                                   & O                                                                   & OM, UT, CA              & Inv                              \\
        6            & Lipton~\etal~\cite{lipton2017precise}              & ICLRw 2017           & DCGAN                        & $\mathcal{Z}$                                                   & O                                                                   & CA                      & Inv                              \\
        7            & PGD-GAN~\cite{PGD-GAN}                             & ICASSP 2018          & DCGAN                        & $\mathcal{Z}$                                                   & O                                                                   & MN, CA                  & Inv                              \\
        8            & Ma~\etal~\cite{ma2018invertibility}                & NeurIPS 2018         & DCGAN                        & $\mathcal{Z}$                                                   & O                                                                   & MN, CA                  & Inv, IP                          \\
        9            & Suzuki~\etal~\cite{suzuki2018spatially}            & ArXiv 2018           & SNGAN, BigGAN, StyleGAN      & $\mathcal{F}$                                                   & 3                                                                   & IN, FL, FF, DA          & CO                               \\
        \rowcolor{green!20}10           & GANDissection~\cite{GANDissection}                 & ICLR 2019            & PGGAN                        & $\mathcal{F}$                                                   & /                                                                   & LS, AD                  & AE, AR                           \\
        11           & NPGD~\cite{NPGD}                                   & ICCV 2019            & DCGAN, SAGAN                 & $\mathcal{Z}$                                                   & L, O                                                                & MN, CA, LS              & Inv, SR, IP                      \\
        \rowcolor{green!20}12           & Image2StyleGAN~\cite{Image2StyleGAN}               & ICCV 2019            & StyleGAN                     & $\mathcal{W}+$                                                  & O                                                                   & FF$^\ddagger$           & Inv, Int, AE, ST                 \\
        13           & Bau~\etal~\cite{bau2019inverting}                  & ICLRw 2019           & PGGAN, WGAN-GP, StyleGAN     & $\mathcal{Z}$, $\mathcal{W}$                                    & L, O                                                                & LS                      & Inv                              \\
        14           & GANPaint~\cite{GANPaint}                           & ToG 2019             & PGGAN                        & $\mathcal{Z}$, $\bm{\Theta}$                                    & L, O, T                                                             & LS                      & Inv, AE                          \\
        \rowcolor{green!20}15           & InterFaceGAN~\cite{InterFaceGAN}                   & CVPR 2020            & PGGAN, StyleGAN              & $\mathcal{Z}$, $\mathcal{W}$                                    & 3, 8                                                                & CH                      & AE, AR                           \\
        \rowcolor{green!20}16           & GANSeeing~\cite{GANSeeing}                         & ICCV 2019            & PGGAN, WGAN-GP, StyleGAN     & $\mathcal{Z}$, $\mathcal{W}$                                    & 13                                                                  & LS                      & Inv                              \\
        17           & YLG~\cite{YLG}                                     & CVPR 2020            & SAGAN                        & $\mathcal{Z}$                                                   & O                                                                   & IN                      & Inv                              \\
        \rowcolor{green!20}18           & Image2StyleGAN++~\cite{Image2StyleGAN++}           & CVPR 2020            & StyleGAN                     & $\mathcal{W}+$, $\mathcal{N}$                                   & O                                                                   & LS, FF                  & Inv, CO, IP, AE, ST              \\
        19           & mGANPrior~\cite{mGANPrior}                         & CVPR 2020            & PGGAN, StyleGAN              & $\mathcal{Z}$                                                   & O                                                                   & FF, CH, LS              & Inv, IC, SR, IP, DN, AE          \\
        20           & MimicGAN~\cite{MimicGAN}                                           & IJCV 2020            & DCGAN                        & $\mathcal{Z}$                                                   & O                                                                   & CA, FF, LF              & Inv, UDA, AD, AN                 \\
        \rowcolor{green!20}21           & PULSE~\cite{PULSE}                                 & CVPR 2020            & StyleGAN                     & $\mathcal{Z}$                                                   & O                                                                   & FF, CH                  & Inv, SR                          \\
        22           & DGP~\cite{DGP}                                     & ECCV 2020            & BigGAN                       & $\mathcal{Z}$                                                   & O, T                                                                & IN, P3                  & Inv, Int, IC, IP, SR, AD, TR, AE \\
        23           & StyleGAN2Distillation~\cite{StyleGAN2Distillation} & ECCV 2020            & StyleGAN2, pix2pixHD         & $\mathcal{W}+$                                                  & /                                                                   & FF                      & AT, AE                           \\
        24           & EditingInStyle~\cite{EditingInStyle}               & CVPR 2020            & PGGAN, StyleGAN, StyleGAN2   & $\mathcal{F}$                                                   & /                                                                   & FF, LS                  & AT                               \\
        25           & StyleRig~\cite{StyleRig}                           & CVPR 2020            & StyleGAN                     & $\mathcal{W}+$                                                  & /                                                                   & FF                      & AT                               \\
        26           & ALAE~\cite{ALAE}                                   & CVPR 2020            & StyleGAN                     & $\mathcal{W}$                                                   & T                                                                   & MN, FF, LS, CH          & Inv, AT                          \\
        \rowcolor{green!20}27           & IDInvert~\cite{IDInvert}                           & ECCV 2020            & StyleGAN                     & $\mathcal{W}+$                                                  & L, O                                                                & FF, LS                  & Inv, Int, AE, CO                 \\
        28           & pix2latent~\cite{pix2latent}                       & ECCV 2020            & BigGAN, StyleGAN2            & $\mathcal{Z}$                                                   & O                                                                   & IN, CO, CF, LS          & Inv, TR, AE                      \\
        29           & IDDistanglement~\cite{IDDisentanglement}           & ToG 2020             & StyleGAN                     & $\mathcal{W}$                                                   & L                                                                   & FF                      & Inv, AT                          \\
        30           & WhenAndHow~\cite{WhenAndHow}                       & ArXiv 2020           & MLP                          & $\mathcal{Z}$                                                   & O                                                                   & MN                      & Inv, IP                          \\
        31           & Guan~\etal~\cite{guan2020collaborative}            & ArXiv 2020           & StyleGAN                     & $\mathcal{W}+$                                                  & L, O                                                                & CH, CD                  & Inv, Int, AT, IC                 \\
        \rowcolor{green!20}32           & SeFa~\cite{SeFa}                                   & CVPR 2021            & PGGAN, BigGAN, StyleGAN      & $\mathcal{Z}$                                                   & 19, 27                                                              & FF, CH, LS, IN, SS, DA  & AE                               \\
        33           & GH-Feat~\cite{GH-Feat}                             & CVPR 2021            & StyleGAN                     & $\mathcal{S}$                                                   & L                                                                   & MN, FF, LS, IN          & Inv, AT, AE                      \\
        \rowcolor{green!20}34           & pSp~\cite{pSp}                                     & CVPR 2021            & StyleGAN2                    & $\mathcal{W}+$                                                  & L                                                                   & FF, AF, CH, CM          & Inv, FF, SI, SR                  \\
        35           & StyleFlow~\cite{StyleFlow}                         & ToG 2021             & StyleGAN, StyleGAN2          & $\mathcal{W}+$                                                  & 12                                                                  & FF, LS                  & AT, AE                           \\
        36           & PIE~\cite{PIE}                                     & ToG 2020             & StyleGAN                     & $\mathcal{W}+$                                                  & O                                                                   & FF                      & AT, AE                           \\
        37           & Bartz~\etal~\cite{bartz2020one}                    & BMVC 2020            & StyleGAN, StyleGAN2          & $\mathcal{Z}$, $\mathcal{W}+$                                   & L                                                                   & FF, LS                  & Inv, DN                          \\
        38           & StyleIntervention~\cite{StyleIntervention}         & ArXiv 2020           & StyleGAN2                    & $\mathcal{S}$                                                   & O                                                                   & FF                      & Inv, AE                          \\
        \rowcolor{green!20}39           & StyleSpace~\cite{StyleSpace}                       & CVPR 2021            & StyleGAN2                    & $\mathcal{S}$                                                   & O                                                                   & FF, LS                  & Inv, AE                          \\
        40           & Hijack-GAN~\cite{Hijack-GAN}                       & CVPR 2021            & PGGAN, StyleGAN              & $\mathcal{Z}$                                                   & /                                                                   & CH                      & AE                               \\
        41           & NaviGAN~\cite{NaviGAN}                             & CVPR 2021            & pix2pixHD, BigGAN, StyleGAN2 & $\bm{\Theta}$                                                   & \cite{StyleGAN2}                                                    & FF, LS, CS, IN          & AE                               \\
        42           & GLEAN~\cite{GLEAN}                                 & CVPR 2021            & StyleGAN                     & $\mathcal{W}+$                                                  & L                                                                   & FF, LS                  & Inv, SR                          \\
        \rowcolor{green!20}43           & ImprovedGANEmbedding~\cite{ImprovedGANEmbedding}   & ArXiv 2020           & StyleGAN, StyleGAN2          & $\mathcal{P}$                                                   & O                                                                   & FF, MF$^\ddagger$       & Inv, IC, IP, SR                  \\
        44           & GFPGAN~\cite{GFPGAN}                               & CVPR 2021            & StyleGAN2                    & $\mathcal{W}$                                                   & L                                                                   & FF                      & Inv, SR                          \\
        45           & EnjoyEditing~\cite{EnjoyEditing}                   & ICLR 2021            & PGGAN, StyleGAN2             & $\mathcal{Z}$                                                   & 12                                                                  & FF, CA, CH, P3, TR      & Inv, AE                          \\
        46           & SAM~\cite{SAM}                                     & ToG 2021             & StyleGAN                     & $\mathcal{W}+$                                                  & L                                                                   & CA, CH                  & AE                               \\
        47           & e4e~\cite{E4E}                                     & ToG 2021             & StyleGAN2                    & $\mathcal{W}+$                                                  & L                                                                   & FF, CH, LS, SC          & Inv, AE                          \\
        \rowcolor{green!20}48           & StyleCLIP~\cite{StyleClip}                         & ICCV 2021            & StyleGAN2                    & $\mathcal{W}+$, $\mathcal{S}$                                   & 47, O                                                               & FF, CH, LS, AF          & AE                               \\
        49           & LatentComposition~\cite{LatentComposition}         & ICLR 2021            & PGGAN, StyleGAN2             & $\mathcal{Z}$                                                   & L                                                                   & FF, CH, LS              & Inv, IP, AT                      \\
        50           & GANEnsembling~\cite{GANEnsembling}                 & CVPR 2021            & StyleGAN2                    & $\mathcal{W}+$                                                  & L, O                                                                & CH, SC, PT              & Inv, AT                          \\
        51           & ReStyle~\cite{ReStyle}                             & ICCV 2021            & StyleGAN2                    & $\mathcal{W}+$                                                  & L                                                                   & FF, CH, SC, LS, AF      & Inv, AE                          \\
        52           & E2Style~\cite{baseline}                            & T-IP 2022            & StyleGAN2                    & $\mathcal{W}+$                                                  & L                                                                   & FF, CH                  & Inv, SI, PI, AT, IP, SR, AE, IH  \\
        \rowcolor{green!20}53           & GPEN~\cite{GPEN}                                   & CVPR 2021            & StyleGAN2                    & $\mathcal{W}+$, $\mathcal{N}$                                   & L                                                                   & FF, CH                  & Inv, SR                          \\
        54           & Consecutive~\cite{Consecutive}                     & ICCV 2021            & StyleGAN                     & $\mathcal{W}+$                                                  & O                                                                   & FF, RA                  & Inv, Int, AE                     \\
        \rowcolor{green!20}55           & BDInvert~\cite{BDInvert}                           & ICCV 2021            & StyleGAN, StyleGAN2          & $\mathcal{F}$/$\mathcal{W}+$                                    & O                                                                   & FF, CH, LS              & Inv, AE                          \\
        56           & HFGI~\cite{HFGI}                                   & CVPR 2022            & StyleGAN2                    & $\mathcal{W}+$, $\mathcal{F}$                                   & L                                                                   & FF, CH, SC              & Inv, AE                          \\
        57           & VisualVocab~\cite{VisualVocab}                     & ICCV 2021            & BigGAN                       & $\mathcal{Z}$                                                   & /                                                                   & P3, IN                  & AE                               \\
        58           & HyperStyle~\cite{HyperStyle}                       & CVPR 2022            & StyleGAN2                    & $\mathcal{W}+$                                                  & L                                                                   & FF, CH, AF              & Inv, AE, ST                      \\
        \rowcolor{green!20}59           & GANGealing~\cite{GANGealing}                       & CVPR 2022            & StyleGAN2                    & $\mathcal{W}$                                                   & /                                                                   & LS, FF, AF, CH, CU      & TR                               \\
        60           & HyperInverter~\cite{HyperInverter}                 & CVPR 2022            & StyleGAN2                    & $\mathcal{W}$, $\bm{\Theta}$                                         & L                                                                   & FF, CH, LS              & Inv, Int, AE                     \\
        \rowcolor{green!20}61           & InsetGAN~\cite{InsetGAN}                           & CVPR 2022            & StyleGAN2                    & $\mathcal{W}+$                                                  & O                                                                   & FF, DF$^\ddagger$       & CO, IG                           \\
        62           & HairMapper~\cite{HairMapper}                       & CVPR 2022            & StyleGAN2                    & $\mathcal{W}+$                                                  & 47                                                                  & FF, CM$^\ddagger$       & AE                               \\
        63           & SAMInv~\cite{SAMInv}                               & CVPR 2022            & BigGAN-deep, StyleGAN2       & $\mathcal{W}+$, $\mathcal{F}$                                   & L                                                                   & FF, LS, IN              & Inv, AE
        \\
        \bottomrule
        \multicolumn{8}{L{1.85\linewidth}}{$^\ast$Abbreviations: AD (\href{https://groups.csail.mit.edu/vision/datasets/ADE20K/}{ADE20K}~\cite{ADE20K}), AF (\href{https://github.com/clovaai/stargan-v2}{AFHQ}~\cite{StarGANv2}), CA (\href{http://mmlab.ie.cuhk.edu.hk/projects/CelebA.html}{CelebA}~\cite{CelebA}), CD (\href{https://bcsiriuschen.github.io/CARC/}{CACD}~\cite{CACD}), CF (\href{http://www.cs.toronto.edu/~kriz/cifar.html}{CIFAR}~\cite{CIFAR}), CH (\href{https://github.com/tkarras/progressive\_growing\_of\_gans}{CelebA-HQ}~\cite{PGGAN}), CM (\href{https://github.com/switchablenorms/CelebAMask-HQ}{CelebAMask-HQ}~\cite{MaskGAN}), CO (\href{https://cocodataset.org/\#home}{MS COCO}~\cite{COCO}), CS (\href{https://www.cityscapes-dataset.com/}{CityScapes}~\cite{Cityscapes}), CU (\href{https://authors.library.caltech.edu/27452/}{Caltech-UCSD Birds}~\cite{CUB}), DA (\href{https://www.gwern.net/Danbooru}{Danbooru}~\cite{Danbooru}, aka Anime Faces), DF (\href{http://mmlab.ie.cuhk.edu.hk/projects/DeepFashion.html}{DeepFashion}~\cite{DeepFashion}), FF (\href{https://github.com/nvlabs/stylegan}{FFHQ}~\cite{StyleGAN}), FL (\href{https://www.robots.ox.ac.uk/~vgg/data/flowers/}{Flowers}~\cite{Flowers}), IN (\href{https://image-net.org/}{ImageNet}~\cite{ImageNet}), LF (\href{http://vis-www.cs.umass.edu/lfw/}{LFW}~\cite{LFW}), LS (\href{https://www.yf.io/p/lsun}{LSUN}~\cite{LSUN}), MF (\href{https://github.com/NVlabs/metfaces-dataset}{MetFaces}~\cite{StyleGAN2-Ada}), MN (\href{http://yann.lecun.com/exdb/mnist/}{MNIST}~\cite{MNIST}), OM (\href{https://github.com/brendenlake/omniglot}{Omniglot}~\cite{Omniglot}), P3 (\href{http://places2.csail.mit.edu/}{Places365}~\cite{Places365}), PL (\href{http://places.csail.mit.edu/}{Places}~\cite{Places}), PT (\href{https://www.robots.ox.ac.uk/~vgg/data/pets/}{Oxford-IIIT Pet}~\cite{CatsAndDogs}, aka Cats and Dogs), RA (\href{https://zenodo.org/record/1188976}{RAVDESS}~\cite{RAVDESS}), SC (\href{http://ai.stanford.edu/~jkrause/cars/car\_dataset.html}{Stanford Cars}~\cite{StanfordCars}), SS (\href{http://streetscore.media.mit.edu/static/files/streetscore\_data.zip}{Streetscape}~\cite{Streetscape}), SV (\href{http://ufldl.stanford.edu/housenumbers/}{SVHN}~\cite{SVHN}), TR (\href{http://transattr.cs.brown.edu/}{Transient}~\cite{Transient}), UT (\href{https://vision.cs.utexas.edu/projects/finegrained/utzap50k/}{UT Zappos50K}~\cite{Shoes})}\\
        \multicolumn{8}{L{1.85\linewidth}}{$^\dagger$Abbreviations: AD (Adversarial Defense), AE (Attribute Editing, \ie, w/o reference), AN (Anomaly Detection), AR (Artifacts Removal), AT (Attribute Transfer, \ie, w/ reference), CO (Image Crossover), [U]DA ([Unsupervised] Domain Adaptation), DN (Image Denoising), FF (Face Frontalization), IC (Image Colorization), IG (Image Generation), IH (Information Hiding), Int (Interpolation), Inv (Inversion), IP (Inpainting), PI (Parsing or Segmentation to Image), SI (Sketch to Image), SR (Image Super-resolution), ST (Style Transfer), TR (Transform and Random Jittering).}\\
        \multicolumn{8}{L{1.85\linewidth}}{$^\ddagger$Some custom datasets collected or regenerated by the authors are omitted since they are not publicly available or can be generated automatically based on current public datasets.}
    \end{tabular}}
\end{table}

\subsection{GAN Inversion}
\label{subsec:GAN_Inversion}
As introduced in \cref{sec:intro}, a natural choice for achieving better output image quality in recent years is to apply adversarial losses, especially the image editing and restoration methods~\cite{Pix2pix,CycleGAN,SRGAN,DeblurGAN}.
However, such a method is insufficient to fully explore the image generation ability of GAN models.
Since the working scheme of GAN models is to map the random vectors (latent representations) into images (\cref{eqn:GAN}), the very first thing to utilize pre-trained GAN models is to invert the input images back into a meaningful latent space (\ie, GAN inversion), then the latent representations could be manipulated or optimized for achieving tasks like image editing or restoration.
Thus we first introduce the commonly used latent spaces in the literature, and then introduce the typical GAN inversion methods.

\subsubsection{Latent Spaces}
\label{subsubsec:latent_space}
In order to invert an image back to the latent representation, the first thing is to determine the inversion target, \ie, the latent space.
For this purpose, the reconstruction accuracy, interpretability, as well as editability should be considered.
In the following, we introduce the commonly used latent spaces in the literature, and an illustration can be found in \cref{fig:GAN_arch}(e).

\vspace{0.5em}
\noindent\textbf{$\mathcal{Z}$ Space \& $\mathcal{C}$ Space}
Following the vanilla GAN~\cite{GAN}, early GAN models~\cite{CGAN,LapGAN,DCGAN,StackGAN,SAGAN,BigGAN,PGGAN} take random noise vectors as the input, and generate fake images via a stack of convolution layers, \ie,
\begin{equation}
    \mathbf{\hat{x}}=\mathit{G}(\mathbf{z},\mathbf{c};\bm{\theta}_\mathit{G}),
    \label{eqn:cGAN}
\end{equation}
where $\mathbf{c}$ is the condition information (\eg, class label, attribute annotations) for conditional GANs~\cite{CGAN,BigGAN}.
Therefore, a natural choice is to directly invert the generation process, and map the image back to a random noise $\mathbf{z}$, which spans the $\mathcal{Z}$ space.
Since $\mathbf{z}$ is sampled from a very simple distribution (\eg, the standard Gaussian distribution), the semantic features are largely entangled in the $\mathcal{Z}$ space, and the simple $\mathcal{Z}$ space is too limited to simultaneously represent both content and semantic information.
An alternative way is to jointly use the $\mathcal{Z}$ space and the $\mathcal{C}$ space, where $\mathbf{c}\in\mathcal{C}$.
Thus some methods (\eg, IcGAN~\cite{IcGAN}) could invert the image into both $\mathcal{Z}$ and $\mathcal{C}$ spaces, showing extraordinary performance in disentangling the content and attribute representations.
However, these methods require massive human efforts in labeling the attributes to model the $\mathcal{Z}$ space.

\vspace{0.5em}
\noindent\textbf{$\mathcal{W}$ Space \& $\mathcal{W}+$ Space}
To separate the semantic conditions (\eg, class, style, \etc) with image contents (\eg, identity), Karras~\etal proposed the StyleGAN series~\cite{StyleGAN,StyleGAN2,StyleGAN3}, which could be formulated by
\begin{equation}
    \mathbf{\hat{x}}=\mathit{G}(\mathbf{m}, \mathbf{n}, \mathit{F}(\mathbf{z}); \bm{\theta}_\mathit{G})=\mathit{G}(\mathbf{m},\mathbf{n},\mathbf{w};\bm{\theta}_\mathit{G}),
    \label{eqn:StyleGAN}
\end{equation}
where $\mathbf{m}$ and $\mathbf{n}$ are learnable constant features and layer-wise noises\footnote{For StyleGAN3~\cite{StyleGAN3}, $\mathbf{m}$ denotes the Fourier feature~\cite{FourierFeature}, while $\mathbf{n}$ is deprecated.}, $\mathit{F}$ is the multi-layer perceptron (MLP) for mapping $\mathbf{z}$ to $\mathbf{w}$.
Due to the superior generation quality of StyleGAN models, they become the most popular GAN architectures.
Instead of mapping back to the $\mathcal{Z}$ space, the latent representation $\mathbf{w}\in\mathcal{W}$ is generally utilized for StyleGAN inversion, which is proven semantically better disentangled with the affine transmissions by the mapping network $\mathit{F}$~\cite{StyleGAN}.
Based on the $\mathcal{W}$ space, some researchers~\cite{Image2StyleGAN} proposed to predict an individual latent representation for each generator layer (see \cref{fig:GAN_arch}(c)), resulting in the $\mathcal{W}+$ space, which shows a better inversion accuracy comparing to the $\mathcal{W}$ space.
Note that the latent codes in the earlier layers (\ie, the ones near the input end) control coarser-grained attributes, while the finer-grained attributes are controlled by the latent codes in the latter layers (\ie, the ones near the output end), and we refer to \cite{StyleGAN} for detailed illustrations.

\vspace{0.5em}
\noindent\textbf{$\mathcal{S}$ Space \& $\mathcal{P}$ Space}
Although $\mathcal{W}$ and $\mathcal{W}+$ spaces have better properties, the changes in $\mathbf{w}$ tend to influence the whole image, and different dimensions in $\mathbf{w}$ follow various distributions.
As introduced in \cref{subsec:Understanding_neuron}, Bau~\etal have explored the relationship between neurons and objects in the generated images.
Inspired by the discovery that each neuron (channel) may be related to a specific class or semantic component, several works choose to predict channel-wise style parameters~\cite{StyleIntervention,StyleSpace,xu2020hierarchical} denoted by $\mathcal{S}$ space, with which one can control the local content of the images.
Besides, the $\mathcal{S}$ space is also extended by considering both channel-wise and spatial-wise diversity~\cite{SalS-GAN}, which is denoted by the $\mathcal{SA}$ space, but we still regard it as a kind of $\mathcal{S}$ space.
Zhu~\etal~\cite{ImprovedGANEmbedding} focused on the distribution of the $\mathcal{W}$ space, and proposed to use the latent representation (denoted by $\mathbf{p}\in\mathcal{P}$) before the last leaky ReLU layer of the mapping network $\mathit{F}$.
Compared to $\mathcal{W}$, $\mathcal{P}$ has a simpler structure (\ie, similar to a multivariate Gaussian distribution).
And they further propose a $\mathcal{P}_\mathit{N}$ space by mapping the distribution of each latent representation to be of zero mean and unit variance.
The $\mathcal{P}_\mathit{N}$ space is better filled by the latent codes, and the image generation quality could be evaluated by the distance between the latent representation and the average latent code.

\vspace{0.5em}
\noindent\textbf{$\mathcal{N}$ Space \& $\mathcal{F}$ Space}
For current GAN models, the largest dimension of the latent representation is $18\times512$ (the 512-dim $\mathbf{w}$ for 18 layers, and the dimension of $\mathbf{z}$ is much smaller).
Even though the most prominent features of the image have been reconstructed, the high-frequency features cannot be faithfully reproduced by such latent spaces.
Therefore, some works~\cite{Image2StyleGAN++,GPEN} proposed to leverage the layer-wise noise maps in StyleGAN and StyleGAN2 (\ie, $\mathbf{n}\in\mathcal{N}$), and achieved much more accurate reconstructions.
Yet the $\mathcal{N}$ space weakens the editability to some extent, thus another category considers the features directly.
For example, Bau~\etal~\cite{GANSeeing} inverted the images into the features at a particular layer to investigate the mode dropping problem.
Kang~\etal~\cite{BDInvert} proposed to map the image back to the feature map $\mathbf{f}$ at a certain scale (\eg, $8\times8$), and jointly use the $\mathcal{F}$ and $\mathcal{W}+$ spaces (denoted by $\mathcal{F}/\mathcal{W}+$ space).

Some methods also consider fine-tuning part of~\cite{GANPaint,NaviGAN} or the whole~\cite{GPEN,NearPerfect} GAN model, and we denote the model parameters as $\bm{\Theta}$ space.
As analyzed in \cref{subsec:Understanding_neuron}, the $\bm{\Theta}$ space is more like the generation rules rather than a latent space, and we mention it for completeness.

\subsubsection{GAN Inversion Methods}

In the literature, there are mainly three kinds of GAN inversion methods, including optimization-based, learning-based, and hybrid methods.
\cref{tab:methods} also summarizes the characteristic of these methods, which shows the used backbone, latent space, inversion method, dataset, and their applications.
In the following, the relevant methods are introduced in detail.

\vspace{0.5em}
\noindent\textbf{Optimization-based Methods}
Given a pre-trained GAN model, the objective of GAN inversion is to find an image $\mathbf{\hat{x}}$ generated by the GAN model which approximates a given image $\mathbf{x}$, \ie,
\begin{equation}
    \mathbf{\hat{x}}^\ast=\arg\min_{\mathbf{\hat{x}}\in\mathcal{X}}\ell(\mathbf{\hat{x}}, \mathbf{x}),
    \label{eqn:inversion_image}
\end{equation}
where $\ell$ is a distance like $\ell_1$ or $\ell_2$, and $\mathcal{X}$ is the image space that are spanned by the images generated by the GAN model.
We require the image is in the GAN space in order to leverage the properties of GANs.
Thus with \cref{eqn:GAN}, it can be rewritten as,
\begin{equation}
    \mathbf{\color{red}z}^\ast=\arg\min_{\color{red}{\mathbf{z}\in\mathcal{Z}}}\ell({\color{red}\mathit{G}(\mathbf{z}, \bm{\delta};\bm{\theta}_\mathit{G})}, \mathbf{x}),
    \label{eqn:inversion_opt}
\end{equation}
where the difference against \cref{eqn:inversion_image} has been highlighted with {\color{red}red}.
Here we use $\mathcal{Z}$ space to illustrate the problem, and the formulation can be generalized to other latent spaces.
Due to that the generator $\mathit{G}$ is differentiable, \cref{eqn:inversion_opt} can be directly optimized via gradient descent algorithms~\cite{creswell2018inverting,lipton2017precise,PGD-GAN,ma2018invertibility,Image2StyleGAN,YLG,Image2StyleGAN++,mGANPrior,MimicGAN,PULSE,DGP,pix2latent,WhenAndHow,PIE,StyleIntervention,StyleSpace,ImprovedGANEmbedding,Consecutive,BDInvert}, and the invertibility of GAN models is discussed by Aberdam~\etal~\cite{WhenAndHow} and Ma~\etal~\cite{ma2018invertibility} with MLP-based and conv-based generators, respectively.

To promote the inversion accuracy and training stability, and boost the subsequent applications such as image editing and restoration, there are many improvements based on the basic formula in \cref{eqn:inversion_opt}.
Zhu~\etal~\cite{zhu2016generative} adopted L-BFGS~\cite{L-BFGS} algorithm instead of Adam~\cite{Adam}.
Lipton~\etal~\cite{lipton2017precise} and Shah~\etal~\cite{PGD-GAN} proposed to use projected gradient descent (PGD) for better optimization in the latent space.
Later, the inner loop of PGD was replaced by a neural network by Raj~\etal~\cite{NPGD}.
Daras~\etal~\cite{YLG} leveraged the attention map of the discriminator to boost the inversion.
Huh~\etal~\cite{pix2latent} and Kang~\etal~\cite{BDInvert} further considered the geometric transformations of the subjects.

An obvious problem of optimization-based methods is efficiency.
Since the optimization procedure usually requires thousands of update iterations, it takes a very long time for inversion.
To eliminate this problem, Abdal~\etal~\cite{Image2StyleGAN} proposed to initialize with the average latent code (\ie, $\mathbf{\bar{w}}$ obtained when training StyleGAN~\cite{StyleGAN}).
However, it still takes even minutes to invert a $1024\times1024$ image with a high-end GPU (\eg, NVIDIA Tesla V100).
An alternative way is to leverage the learning-based methods, which predict the latent representation in a single forward pass.

\vspace{0.5em}
\noindent\textbf{Learning-based Methods}
Learning-based GAN inversion methods~\cite{IcGAN,IDDisentanglement,GH-Feat,pSp,StyleFlow,bartz2020one,GLEAN,GFPGAN,SAM,E4E,LatentComposition,ReStyle,baseline,GPEN} are generally equipped with an additional encoder $\mathit{E}$, and the latent code is optimized in an indirect way.
Specifically, they optimize the parameters of the encoder, which is used to map the image back to the latent code, \ie, $\mathit{E}:\mathbf{x}\to\mathbf{z}$.
Therefore, \cref{eqn:inversion_opt} can be re-written as,
\begin{equation}
    {\color{red}\bm{\theta}}_\mathit{\color{red}E}^\ast=\arg\min_{\color{red}\bm{\theta}_\mathit{E}}{\color{red}{\sum}_\mathit{i}}\ell(\mathit{G}({\color{red}\mathit{E}(\mathbf{x}_\mathit{i};\bm{\theta}_\mathit{E})}, \bm{\delta};\bm{\theta}_\mathit{G}), \mathbf{x}_{\color{red}\mathit{i}}),
    \label{eqn:inversion_learning}
\end{equation}
where $\bm{\theta}_\mathit{E}$ denotes the parameters of $\mathit{E}$, and the difference against \cref{eqn:inversion_opt} has been highlighted in {\color{red}red}.

It can be seen that the learning-based methods train the encoder with a whole dataset to learn the mapping from image to latent code, rather than optimize on a single image.
Such a scheme yields several advantages.
First, the optimization becomes smoother, which avoids the latent code from being trapped into a local optimum.
Then, since the encoder has access to a large batch of images, it is easier for the encoder to learn some patterns from the data, \eg, the semantic directions for image manipulation in the latent space~\cite{IcGAN,IDDisentanglement,GH-Feat,StyleFlow,E4E,LatentComposition}.
On the contrary, the optimization-based methods generally rely on other methods (will be introduced in \cref{sec:disentanglement}).
Moreover, while the optimization-based methods can only be optimized with real images (the latent code of the synthetic images are already available), the learning-based methods can be trained with synthetic images.
In this way, they directly leverage the ground-truth latent code~\cite{IcGAN}, and \cref{eqn:inversion_learning} can be written as,
\begin{equation}
    \bm{\theta}_\mathit{E}^\ast=\arg\min_{\bm{\theta}_\mathit{E}}{\sum}_\mathit{i}\ell({\color{red}\mathit{E}}({\color{red}\mathit{G}}({\color{red}\mathbf{z}}_\mathit{i}, \bm{\delta};\bm{\theta}_\mathit{G}); \bm{\theta}_\mathit{E}), {\color{red}\mathbf{z}}_\mathit{i}).
    \label{eqn:inversion_learning_latent_code}
\end{equation}
Some methods also introduce extra models as auxiliary information or guidance.
For example,
Nitzan~\etal~\cite{IDDisentanglement} incorporated a face landmark detection model to control the pose of the generated face images.
Many works~\cite{IDDisentanglement,pSp,GPEN} introduced face recognition~\cite{ArcFace,CurricularFace} or contrastive learning~\cite{MoCo} models as ID/content similarity metrics, while some of them~\cite{baseline,GANDissection} also introduced semantic segmentation or face parsing models.
Furthermore, vision-language models are also used for inversion and editing~\cite{StyleClip}.

With the encoder, the model can get the latent code in a single forward pass, but the inversion is generally less accurate than optimization-based methods.
Therefore, Wei~\etal~\cite{baseline} proposed to use a multi-stage refinement scheme to keep both accuracy and efficiency, Alaluf~\etal~\cite{ReStyle} introduced an iterative procedure which is initialized with the average face latent $\mathbf{\bar{w}}$.

\begin{figure}
    \centering
    \scriptsize
    \begin{overpic}[width=.99\linewidth]{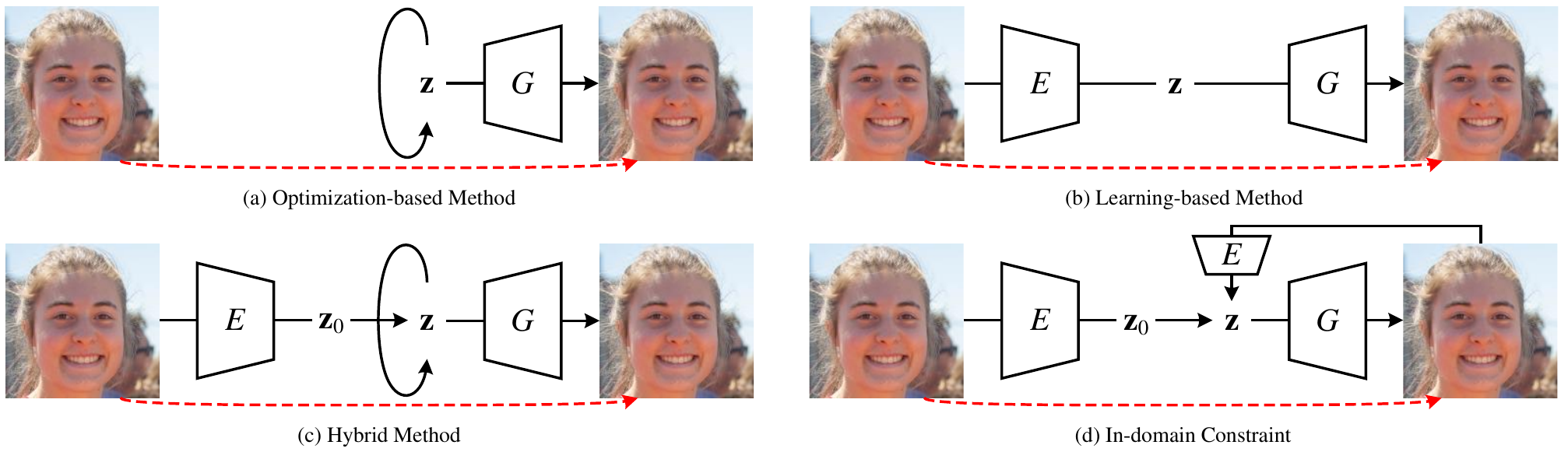}
        \put(5, 17){$\mathbf{x}$}
        \put(43, 17){$\mathbf{\hat{x}}$}
        \put(56, 17){$\mathbf{x}$}
        \put(94, 17){$\mathbf{\hat{x}}$}

        \put(5, 2){$\mathbf{x}$}
        \put(43, 2){$\mathbf{\hat{x}}$}
        \put(56, 2){$\mathbf{x}$}
        \put(94, 2){$\mathbf{\hat{x}}$}
    \end{overpic}
    \caption{Illustration of GAN inversion methods. $\mathbf{x}$ and $\mathbf{\hat{x}}$ are given real image and generated image, respectively. The {\color{red}red} dotted line means supervision. It can be seen that the in-domain constraint~\cite{IDInvert} requires the generated image $\mathbf{\hat{x}}$ can be inverted back into the latent space. Here, $\mathbf{z}$ is not restricted in $\mathcal{Z}$ space, and may refer to more generic latent code (\eg, $\mathbf{w}$, $\mathbf{f}$, \etc).}
    \label{fig:inversion_scheme}
\end{figure}
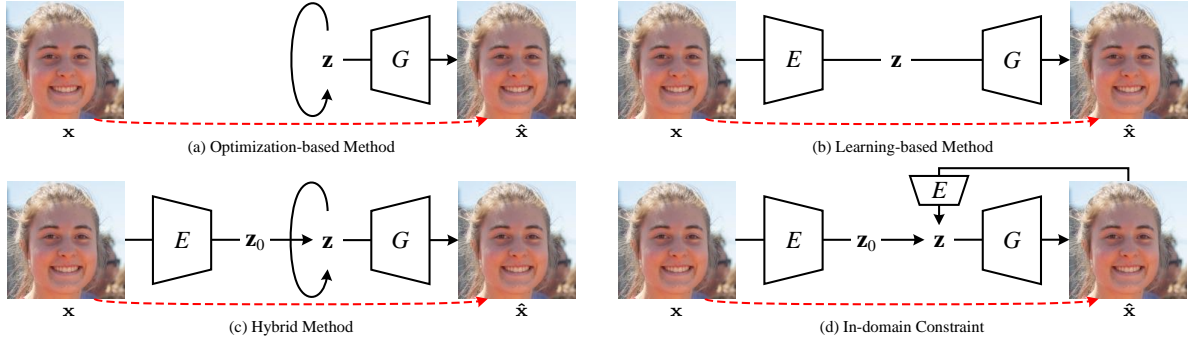

\vspace{0.5em}
\noindent\textbf{Hybrid Methods}
For both efficiency and accuracy, Zhu~\etal~\cite{zhu2016generative} proposed the hybrid method that initializes the latent code for optimization via learning-based methods, which was followed by a series of works~\cite{NPGD,GANPaint,IDInvert,guan2020collaborative,GANEnsembling}.
Among these methods, we would like to highlight the method named IDInvert~\cite{IDInvert}, which leverages the encoder for in-domain inversion.
Specifically, an encoder is trained like the learning-based methods following \cref{eqn:inversion_learning}, and the discriminator is also leveraged for training.
Then, following the optimization-based methods, the latent code $\mathbf{z}_0$ obtained by the encoder is used to initialize the optimization procedure, and then the learning objective can be formulated by,
\begin{equation}
    \mathbf{z}^\ast=\argmin_{\mathbf{z}\in\mathcal{Z}}\ell(\mathit{G}(\mathbf{z}, \bm{\delta};\bm{\theta}_\mathit{G}), \mathbf{x}) {\color{red}+ \lambda\|\mathbf{z}-\mathit{E}(\mathit{G}(\mathbf{z}, \bm{\delta};\bm{\theta}_\mathit{G}); \bm{\theta}_\mathit{E})\|_2},
    \label{eqn:inversion_idinvert}
\end{equation}
where $\lambda$ is the balancing hyper-parameter, and the difference between \cref{eqn:inversion_opt} has been highlighted with {\color{red}red}.
It can be seen from \cref{eqn:inversion_idinvert}, that the second term requires that the image generated with the optimized latent code can be mapped back via the encoder.
In other words, the optimization is guaranteed to be performed within the manifold supported by the encoder (as well as the generator), and it is called \textit{in-domain inversion}.
Please refer to \cref{fig:inversion_scheme}(d) for an illustration.

Guan~\etal~\cite{guan2020collaborative} leveraged the encoder during the optimization process in a collaborative way.
In particular, the encoder (\ie, the embedding network in their paper) and the iterator participate in each other's procedure.
In each iteration, the encoder predicts a latent code as a reasonable initialization for the iterator, which obviously speeds up the optimization.
Then, the iterator obtains a better latent code via optimization, which conversely serves as supervision for training the encoder.
Both the encoder and the iterator achieve better performance/efficiency, and the method naturally supports online learning.

Apart from the above methods, some works choose to learn an encoder together with the generator~\cite{BiGAN,ALI,ALAE}, which combine autoencoders with GANs and are out of the scope of this review.
Some other generative models are invertible by design~\cite{GLOW,DDPM}, however, the latent representation is typically a tensor of the same size as the image, which is sampled from a very simple distribution (\eg, Gaussian), making the latent representation less meaningful, and is beyond the scope of this paper.

\subsection{Semantic Disentanglement and Discovery}
\label{sec:disentanglement}

Understanding visual concepts is vital for many vision tasks like image editing and restoration, which helps the models to generate semantically consistent results.
In this section, we review the relevant methods from two perspectives, \ie, semantic disentanglement and discovery~\cite{ramesh2018spectral}.
The former cares more about encouraging the independence of latent variables, \ie, a perturbation to any component of a latent variable should result in a change to the output that is distinct, while the latter focuses on discovering disentangled latent representations of highly interpretability, \ie, a perturbation to one dimension should result in a change along exactly one meaningful attribute.

\subsubsection{Semantic Disentanglement}
\label{subsubsec:semantic_disentanglement}

In early GAN models (\eg, DCGAN~\cite{DCGAN}), the concepts or attributes are generally entangled to a large extent in the rather simple latent space (\eg, standard Gaussian).
Although some methods~\cite{IcGAN} achieve disentangled attributes by leveraging the conditional GANs~\cite{CGAN,BigGAN} trained on datasets with class labels (\eg, ImageNet~\cite{ImageNet}, LSUN~\cite{LSUN}, \etc) or attribute annotations (\eg, CelebA~\cite{CelebA}, RaFD~\cite{RaFD}, \etc), massive human efforts are required, which is unacceptable for all tasks.
Thus a series of methods~\cite{InfoGAN,SpectralRegularizer,HessianPenalty,VP-GAN,WhereAndWhat,OroJaR,EigenGAN} are proposed for better disentanglement in GANs.

A route of methods achieves disentanglement by maximizing the mutual information.
InfoGAN~\cite{InfoGAN} is an early exploration on this task, which introduces an extra latent code $\mathbf{c}$ alongside the original random variable $\mathbf{z}$.
The model architecture is the same as \cref{eqn:cGAN}, while the meaning of $\mathbf{c}$ is automatically learned via mutual information maximization when optimizing the GAN models, \ie,
\begin{equation}
    \min_\mathit{G}\max_\mathit{D}\mathcal{L}_\mathit{InfoGAN}(\mathit{G},\mathit{D})=\mathcal{L}_\mathit{GAN}(\mathit{G},\mathit{D})-\lambda\mathit{I}(\mathit{G}(\mathbf{z},\mathbf{c};\bm{\theta}_\mathit{G}); \mathbf{c}),
    \label{eqn:loss_infogan}
\end{equation}
where $\mathit{I}(\mathit{x};\mathit{y})$ denotes the mutual information between $\mathit{x}$ and $\mathit{y}$, and $\mathcal{L}_\mathit{GAN}$ is defined in \cref{eqn:loss_GAN}.
In particular, a regressor is trained to predict the input $\mathbf{c}$.
As such, the changes in $\mathbf{c}$ are required to influence the generation in a distinct way, which forces $\mathbf{c}$ to be traceable and disentangled.
Zhu~\etal~\cite{VP-GAN} argued that the latent variation predictability can represent the disentanglement, and improved InfoGAN by predicting the relative changes between two generated images with only one dimension of the latent vector changed, \ie,
\begin{equation}
    \min_\mathit{G}\max_\mathit{D}\mathcal{L}_\mathit{VP-GAN}(\mathit{G},\mathit{D})=\mathcal{L}_\mathit{GAN}(\mathit{G},\mathit{D})-\lambda\mathit{I}(\mathit{G}(\mathbf{z};\bm{\theta}_\mathit{G}), \mathit{G}(\mathbf{z}+ \mathbf{\epsilon}\mathbf{d};\bm{\theta}_\mathit{G}); \mathbf{d}),
    \label{eqn:loss_vp_gan}
\end{equation}
where $\mathbf{d}$ is a one-hot vector with the same dimension as $\mathbf{z}$, and $\epsilon\!\in\!\mathcal{N}(0, 1)$.
Note that the first metric of GAN disentanglement is proposed based on the observation.
They further assumed that the interpretable representation should be spatially consistent, and improved the variation predictability in \cite{WhereAndWhat}.

Another category focuses on the gradient of the GAN models with respect to the inputs.
Ramesh~\etal~\cite{SpectralRegularizer} focused on the coordinate directions in the latent space, and proposed a spectral regularizer to align the leading right-singular vectors of the Jacobian of GAN models \wrt the inputs with the coordinate axes, so that the trajectories are semantically meaningful.
Peebles~\etal~\cite{HessianPenalty} achieved disentanglement by encouraging the Hessian matrix of the GAN models \wrt the inputs to be diagonal, and the Hutchinson's estimator is leveraged to approximate the regularization.
It is worth noting that the authors also showed the ability of the Hessian Penalty to ``deactivate'' redundant components.
Wei~\etal~\cite{OroJaR} required that when perturbing a single dimension of the network input, the change in the output should be independent (and uncorrelated) with those caused by the other input dimensions.
Based on this assumption, they changed the assumption in \cite{HessianPenalty} from $\frac{\partial}{\partial\mathit{z}_\mathit{j}}(\frac{\partial\mathit{G}}{\partial\mathit{z}_\mathit{i}})=0$ to $[\frac{\partial\mathit{G}}{\partial\mathit{z}_\mathit{j}}]^\mathsf{T}\frac{\partial\mathit{G}}{\partial\mathit{z}_\mathit{i}}=0$.
He~\etal~\cite{EigenGAN} embedded a linear subspace in each layer of the GAN model by directly adding a group of basis $\mathbf{U}_\mathit{i}$, where the $\mathit{i}$-th noise $\mathbf{z}_\mathit{i}$ is rewritten as $\mathbf{U}_\mathit{i}\mathbf{L}_\mathit{i}\mathbf{z}_\mathit{i}+\bm{\mu}_\mathit{i}$, and $\mathbf{L}_\mathit{i}$ is a diagonal matrix indicating the importance of basis vectors, $\bm{\mu}_\mathit{i}$ denotes the origin of the subspace.
They required that the basis is orthogonal, \ie, minimizing $\|\mathbf{U}_\mathit{i}^\mathsf{T}\mathbf{U}_\mathit{i}-\mathbf{I}\|^2_\mathit{F}$.

As introduced in \cref{sec:GANs_development}, recent large-scale GAN models (especially the StyleGAN~\cite{StyleGAN} series) achieve disentangled latent representations by sending the conditional information to the generator layers (see \cref{fig:GAN_arch}), and the StyleGAN model maps the latent code into an intermediate latent space, \ie, $\mathcal{W}$ space via an MLP.
A series of regularization terms are deployed for more stable training and better disentanglement, \eg, the mixing regularization and perceptual path length regularization.
Please refer to \cref{tab:GAN_comparison} for more information.
However, since the disentanglement is implicitly achieved in the $\mathcal{W}$ space, researchers also focused on finding meaningful semantic directions in the GAN models, which are introduced as follows.

\subsubsection{Semantic Discovery}

To find meaningful semantic directions in the latent spaces, there are two main streams in the literature, \ie, the supervised and unsupervised methods, which are introduced respectively as follows.

\vspace{0.5em}
\noindent\textbf{Supervised Methods}
The most intuitive way to leverage the manually annotated supervision is to model the semantic space explicitly via conditional GANs, where the attribute vector is delivered into the GAN model.
Perarnau~\etal~\cite{IcGAN} followed this intuition in IcGAN and adopted two encoders to map images into both the random noise space and the attribute space.
However, this scheme is unavailable for unconditional GANs, which only synthesize images from random noise.
Besides, the methods require massive human efforts for data annotation, which makes it inflexible and expensive.

Therefore, researchers sought to find indirect ways to leverage the annotations.
For example, Shen~\etal~\cite{InterFaceGAN} assumed that there is a hyperplane in the latent space for binary attributes (\eg, male and female), which serves as the separation boundary between the opposite attributes.
Thus they proposed to leverage the \textit{normal vector of the hyper-plane} as the editing direction, and the support vector machine (SVM) is deployed for obtaining the hyper-plane.
\cite{Hijack-GAN,StyleGAN2Distillation, EnjoyEditing} followed a similar assumption.
Wang~\etal~\cite{Hijack-GAN} trained a proxy model which can map the input noise to the attribute space, and the \textit{gradient} of the proxy model \wrt the input noise is used as the non-linear editing direction.
Viazovetskyi~\etal~\cite{StyleGAN2Distillation} leveraged the pre-trained image attribute classifier to find the \textit{class centers} of attributes in the latent space, and used the direction between different centers to represent the editing direction.
Zhuang~\etal~\cite{EnjoyEditing} introduced a group of learnable \textit{editing directions} $\mathbf{d}$, and the pre-trained image attribute classifier is used to align each of the directions with an attribute.

Instead of learning editing directions, Abdal~\etal~\cite{StyleFlow} introduced a conditional normalizing flow model in $\mathcal{W}$-space of StyleGAN, which takes the attributes as conditions to map the latent $\mathbf{w}$ back into the noise space.
To edit the attributes, one can obtain the edited latent $\mathbf{w}'$ by inverting the noise with new attributes.
Besides, several methods focused on editing specific attributes in similar ways, for example, Alaluf~\etal~\cite{SAM} sought to edit the \textit{age} via an age regressor, and Goetschalckx~\etal~\cite{GANalyze} tried to make the images \textit{more memorable} with a memorability predictor.

The methods mentioned above still find the attribute directions directly related to the supervision (\ie, the manually annotated attributes), and some researchers proposed to edit some attributes with indirect supervision.
As introduced in \cref{subsec:Understanding_neuron}, Bau~\etal~\cite{GANDissection} have revealed the relationship between the GAN model neurons and the appearance of objects, which inspires a series of disentanglement methods to achieve \textit{spatially precise control}~\cite{GH-Feat,StyleSpace,StyleIntervention}.
Nitzan~\etal~\cite{IDDisentanglement} leveraged a pre-trained \textit{identification} model and a \textit{landmark detection} model, with which the attributes that can be described via landmarks (\eg, expression) are decomposed with the identity, thus the attributes can be transferred to other faces.
Tewari~\etal~\cite{StyleRig} shared a similar idea but employed a pre-trained three-dimensional morphable face model (\textit{3DMM}) to learn synthetic face image editing, which is later extended to real faces in \cite{PIE}.
Huh~\etal~\cite{pix2latent} learned transformation of input shift, rescaling, \etc, and estimate spatial transformation together with the latent variable.
In particular, the spatial positions of the objects are obtained via a pre-trained \textit{object detection} model.
Xu~\etal~\cite{Consecutive} bypassed the requirement of the pre-trained classifier by leveraging video data and optical flow.
Jahanian~\etal~\cite{Steerability} applied data augmentation methods to the original data and train the model in a self-supervised manner, but they are limited to these simple semantic features.


\vspace{0.5em}
\noindent\textbf{Unsupervised Methods}
Compared to supervised methods, the unsupervised scheme has two main advantages.
First, it avoids the need for manual labeling, thus is more applicable to generalize to other categories, which greatly promotes the practical value of the methods.
Second, the lack of supervision information also implies the lifting of restrictions, \ie, the models are encouraged to find new semantic features, which are not necessarily labeled by human annotators.

Some methods for semantic disentanglement introduced in \cref{subsubsec:semantic_disentanglement} also support semantic discovery, which mainly considers the gradient \wrt inputs or mutual information.
For example, Ramesh~\etal~\cite{SpectralRegularizer} found trajectories corresponding to the \textit{principal eigenvectors}, which is extended by Wang~\etal~\cite{wang2021geometric} via further introducing the \textit{Riemannian geometry metric}.
Peebles~\etal~\cite{HessianPenalty} leveraged their \textit{Hessian-based regularization} term to identify interpretable directions in BigGAN's latent space.
Voynov~\etal~\cite{voynov2020unsupervised} adopted similar strategies as \cite{VP-GAN}, which learns a group of editing basis.
By adding a perturbation to the latent code, the model is required to \textit{predict the perturbation} given the original and modified image pairs.
Tzelepis~\etal~\cite{WarpedGANSpace} further argued that existing methods assume the latent directions are linear, which limits the disentanglement, thus they introduce RBF kernel to map the latent representations into a \textit{non-linear space}, and learn a non-linear RBF path.

As analyzed in \cref{subsubsec:GANDissection}, the final output is determined by the features in the previous layers, thus GAN features can be considered for easier semantic discovery than the image domain.
Following this idea, H{\"a}rk{\"o}nen~\etal~\cite{GANSpace} proposed that the \textit{principal components of early layer features} represent important factors of variations, so that they find semantic directions by PCA operation over the early feature space.
Collins~\etal~\cite{EditingInStyle} sought to find attributes relevant to a region of interest (ROI), and proposed to cluster the features in a layer and sort the channels by the \textit{relevance with the ROI regions} (following the idea of \cite{GANDissection} but in an unsupervised way).
From another perspective, the parameters also play an important role in the generation process, which has also been introduced in \cref{subsubsec:GANrewriting}.
Therefore, there are also some methods leveraging the parameter space for semantic discovery.
Shen and Zhou~\cite{SeFa} formulated the perturbation in a layer and the corresponding changes in the output image, and finally the relationship between the semantic directions and the parameter matrix is derived, leading to a \textit{closed-form solution}, which achieves an efficient and effective semantic factorization method (termed by SeFa).
Cherepkov~\etal~\cite{NaviGAN} followed similar strategies to \cite{voynov2020unsupervised} and \cite{wang2021geometric}, and proposed two methods in the parameter domain by considering parameter perturbation and the Hessian of LPIPS \wrt the GAN parameters.

It is worth noting that the large-scale vision-language models are introduced by Patashnik~\etal~\cite{StyleClip}, which leverages the common latent space for visual-linguist features, and combines the image-domain semantic discovery with the intrinsic semantic information in languages.

\section{Applications to Image Editing and Restoration}
\label{sec:applications}

Since most GAN models are trained in an unsupervised manner, they are not limited to a specific task and can serve as a generic prior for many other tasks with visual outputs (\eg, images).
In this section, we will go through these applications, where a brief summary can be found in \cref{tab:methods}.

\subsection{Image Editing}
\label{subsec:app_image_editing}

As introduced in \cref{subsec:GAN_Inversion,sec:disentanglement}, the latent spaces of recent GAN models have been widely explored in the literature.
With the exhilarating attribute disentanglement ability of the latest GAN models, we are able to discover and traverse the latent spaces in a more semantic-aware way, making image editing applicable with pre-trained GAN models.
In this section, we briefly review the image editing applications relying on semantic disentanglement and discovery methods, including image interpolation, style transfer, image crossover, attribute editing, image transform, \etc.

\subsubsection{Image Interpolation (Image Morphing)}

Image interpolation, also known as image morphing, aims at interpolating two images semantically.
By observing the generated images, we can have an intuitive sense of the space used for interpolation, \ie, a well-characterized space should be reflected in smooth and semantically appropriate transitions in the interpolated images.
Therefore, apart from generating new images, image interpolation is often utilized to evaluate the quality of the discovered latent space.
For example, as shown in \cref{fig:intro_interpolation}, the interpolation operation in the pixel space (\ie, $\mathbf{I}=\alpha\cdot\mathbf{x}_1+(1-\alpha)\cdot\mathbf{x}_2$) leads to obvious artifacts, and the results cannot meet the semantic expectations.
On the contrary, the interpolation in the latent space (\eg, \cite{zhu2016generative,Image2StyleGAN,Consecutive,IDInvert}) is able to well describe the change of view, which can be formulated by,
\begin{equation}
    \mathbf{I}=\mathit{G}(\beta\cdot\mathbf{z}_1+(1-\beta)\cdot\mathbf{z}_2); \bm{\theta}_\mathit{G}),
    \label{eqn:interpolation}
\end{equation}
where $\mathbf{z}_1$ and $\mathbf{z}_2$ can be obtained by the inversion methods (\cref{subsec:GAN_Inversion}) from two images $\mathbf{x}_1$ and $\mathbf{x}_2$.

\begin{figure}[t]
    \centering
    \scriptsize
    \begin{tabular}{cccccc}
        \hspace{-2.75mm} \includegraphics[width=.15\linewidth]{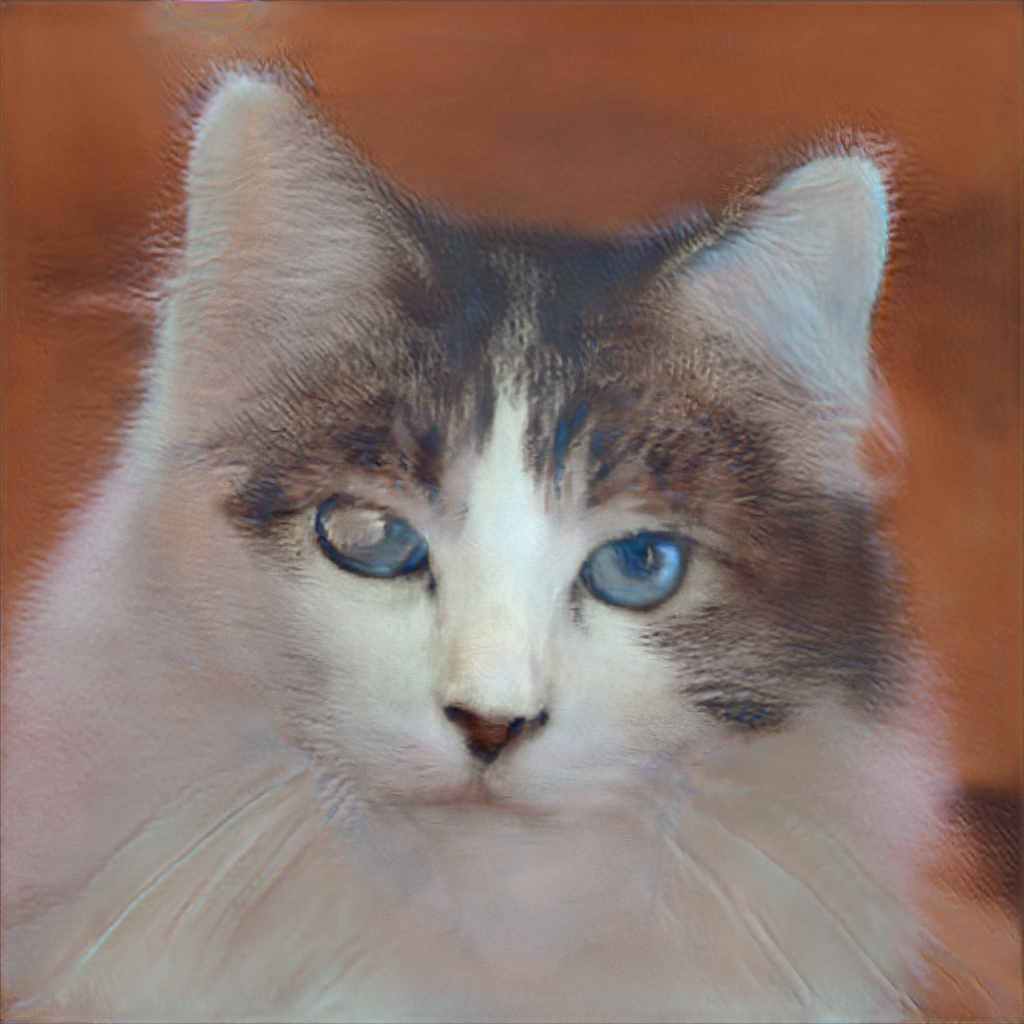}  \hspace{-2.75mm} & \hspace{-2.75mm} \includegraphics[width=.15\linewidth]{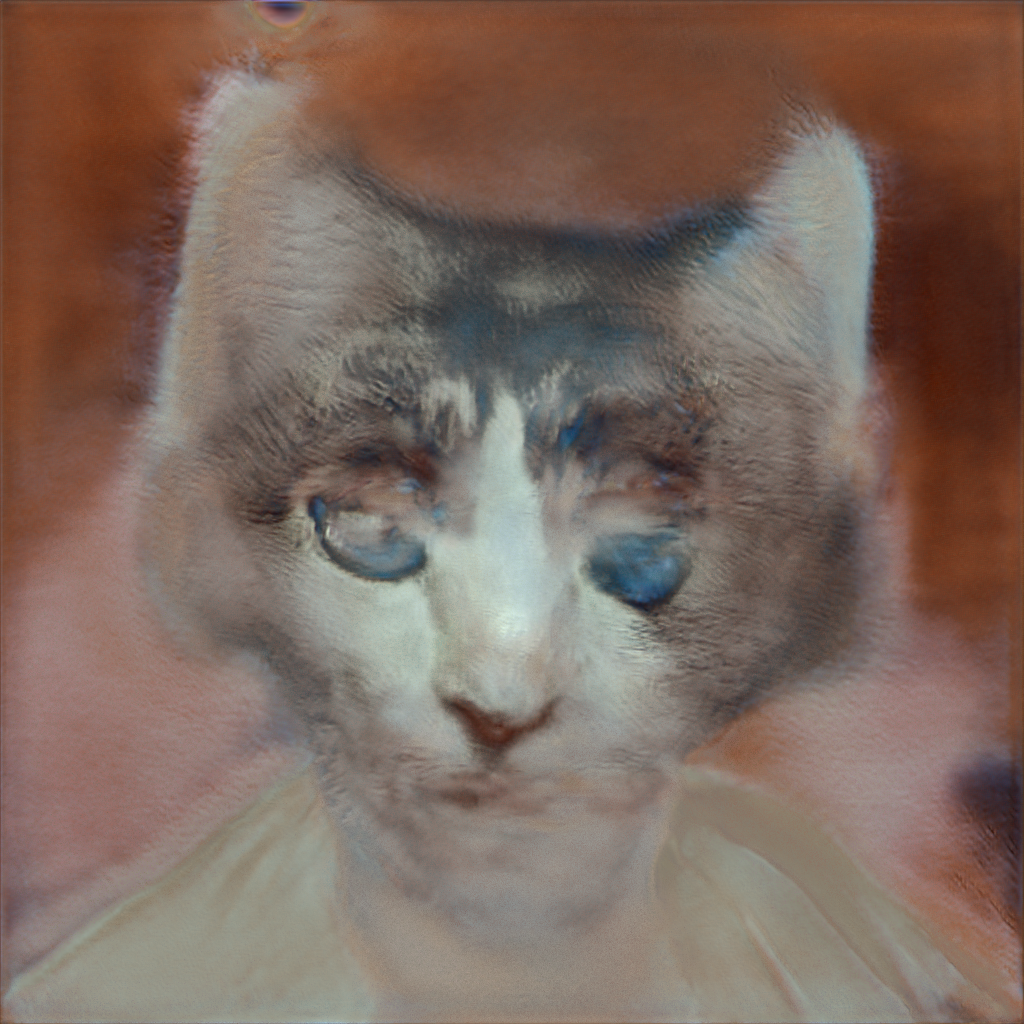}  \hspace{-2.75mm} & \hspace{-2.75mm} \includegraphics[width=.15\linewidth]{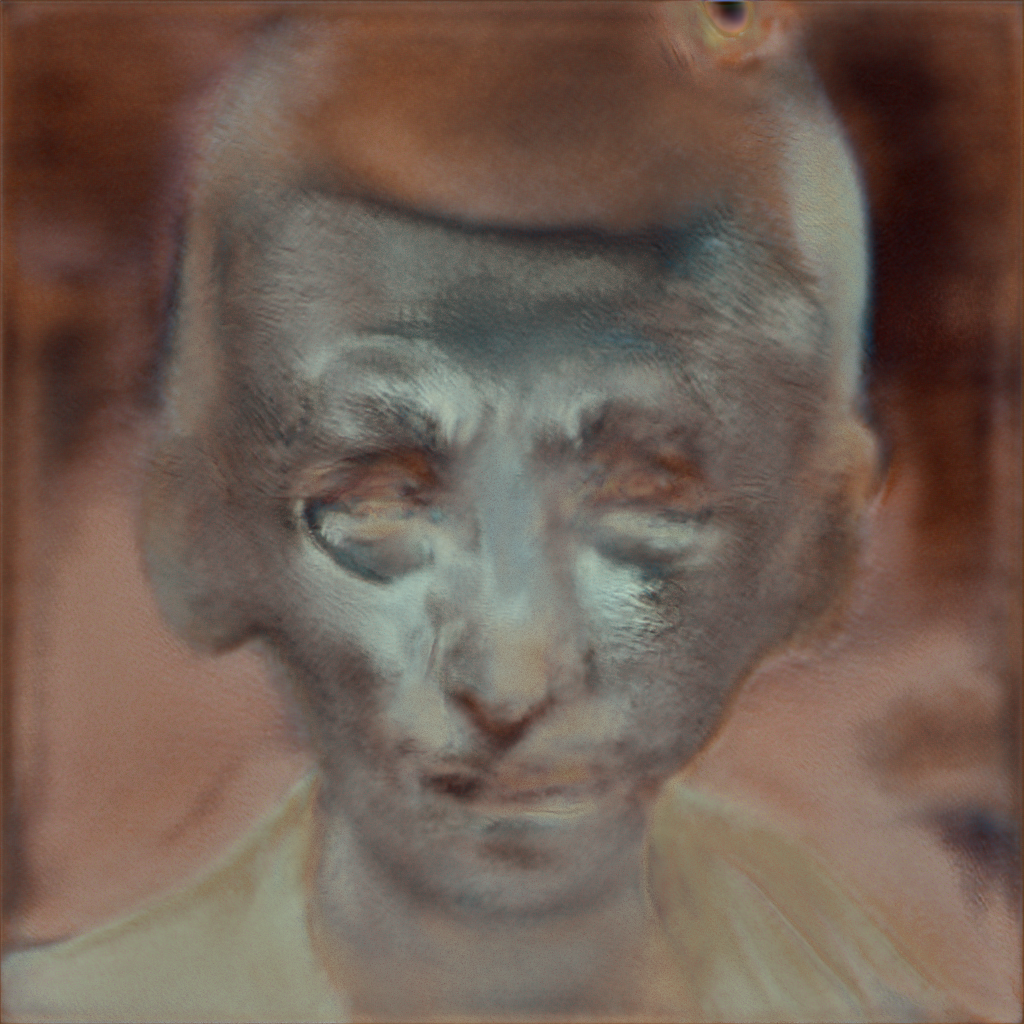}  \hspace{-2.75mm} & \hspace{-2.75mm} \includegraphics[width=.15\linewidth]{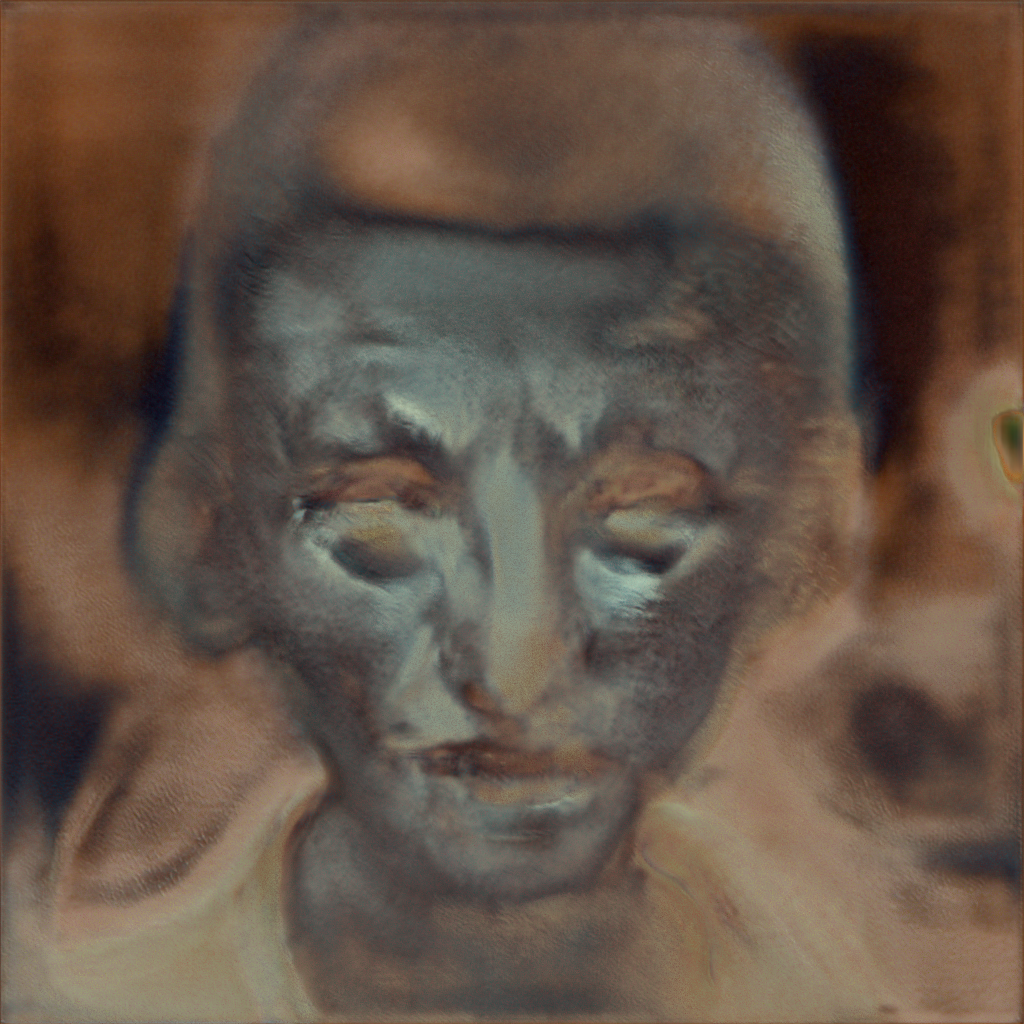}  \hspace{-2.75mm} & \hspace{-2.75mm} \includegraphics[width=.15\linewidth]{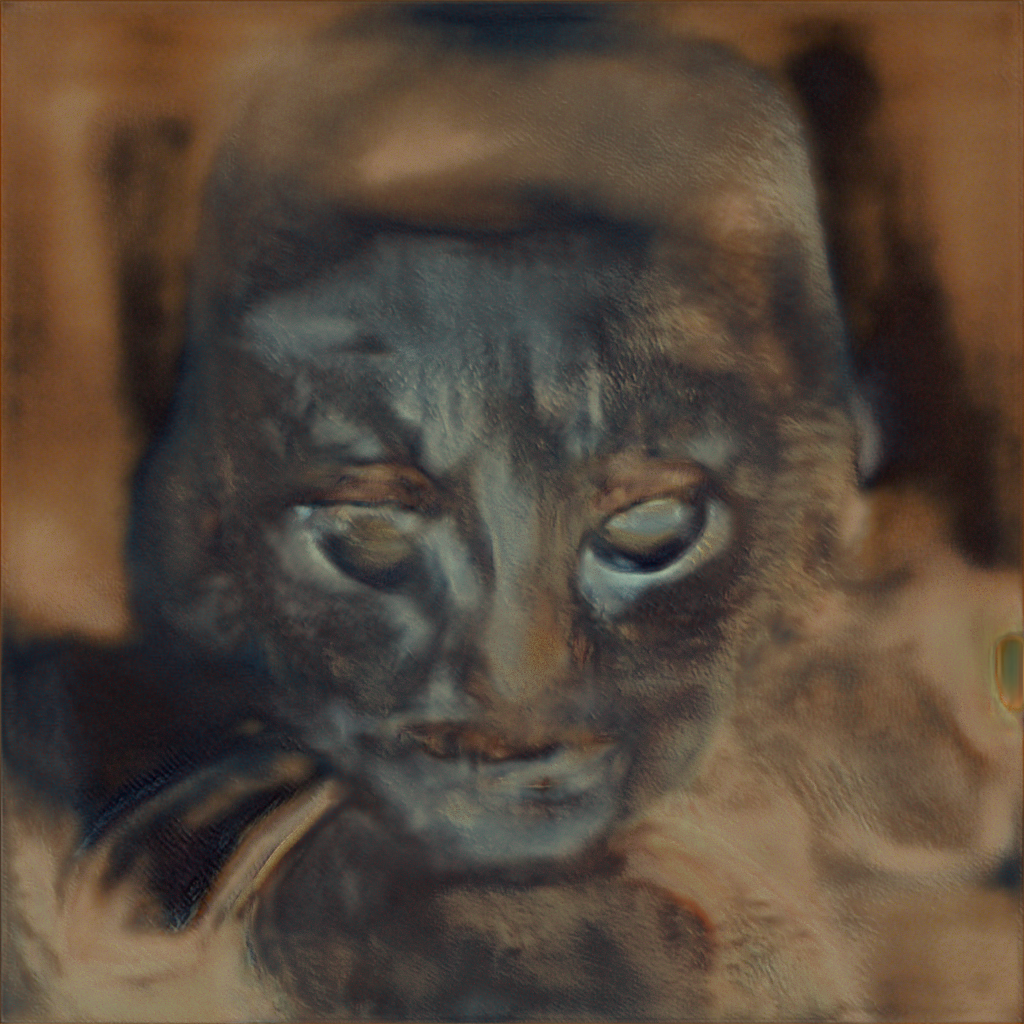}  \hspace{-2.75mm} & \hspace{-2.75mm} \includegraphics[width=.15\linewidth]{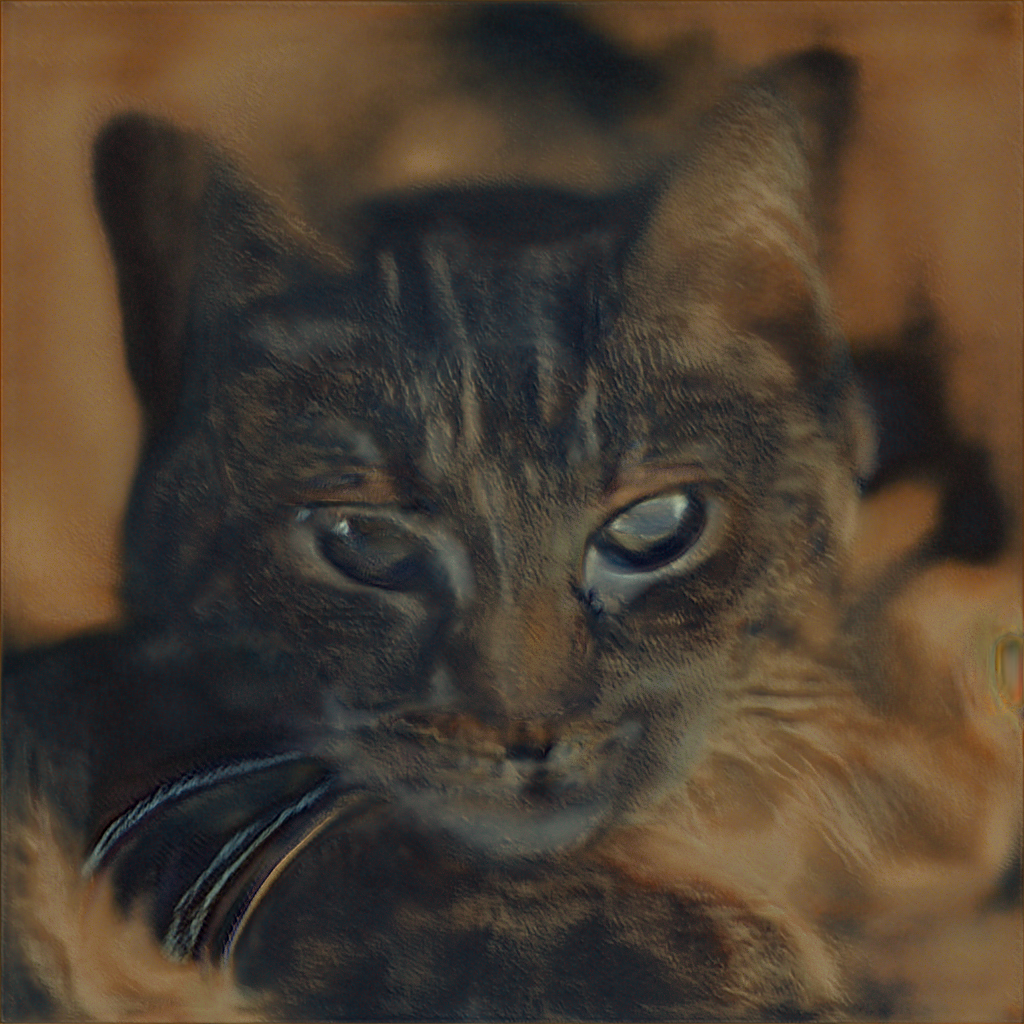}\hspace{-2.75mm}  \\[-1mm]
        \hspace{-2.75mm} $\mathit{G}(\mathbf{z}_1)$  \hspace{-2.75mm} & \hspace{-2.75mm} $\lambda=$0.2  \hspace{-2.75mm} & \hspace{-2.75mm} $\lambda=$0.4  \hspace{-2.75mm} & \hspace{-2.75mm} $\lambda=$0.6  \hspace{-2.75mm} & \hspace{-2.75mm} $\lambda=$0.8  \hspace{-2.75mm} & \hspace{-2.75mm} $\mathit{G}(\mathbf{z}_2)$ \hspace{-2.75mm}
    \end{tabular}
    \caption{Inversion and interpolation~\cite{Image2StyleGAN} between two cat images A and B from the AFHQ dataset~\cite{StarGANv2} with a GAN model trained on the FFHQ dataset~\cite{StyleGAN}. It can be seen that even the image can be well reconstructed, the interpolation results ($\mathit{G}(\lambda\cdot\mathbf{z}_1+(1-\lambda)\cdot\mathbf{z}_2)$) are looming the shape of human faces. $\mathbf{z}_1$ and $\mathbf{z}_2$ are latent codes obtained from $\mathbf{x}_1$ and $\mathbf{x}_2$ via \cite{Image2StyleGAN}, respectively.}
    \label{fig:interpolation}
\end{figure}

There are two special cases that are worth mentioning.
As revealed by Abdal~\etal~\cite{Image2StyleGAN}, when performing GAN inversion via optimization-based methods, a GAN model trained with face datasets can well reconstruct an image from other classes (\eg, cats).
However, since the optimization-based methods are less constrained and can have a much larger accessible latent space, interpolating between these reconstructed images will not produce meaningful results (see \cref{fig:interpolation}), which further verifies the propositions of IDInvert~\cite{IDInvert} about domain-regularized optimization (see \cref{eqn:inversion_idinvert}).
Another thing is about the training/fine-tuning based inversion methods (see \cref{tab:methods}).
For more accurate reconstruction, some of these methods (\eg, \cite{DGP}) will derive a specific group of model parameters for each sample.
In this way, there will be two sets of $\bm{\theta}_\mathit{G}$ when performing image interpolation operations, which is inconsistent with \cref{eqn:interpolation}.
Fortunately, Wang~\etal~\cite{wang2019deep} showed that when fine-tuned from the same checkpoint, the parameters of two models can be interpolated to achieve an intermediate function.
Thus, the image interpolation can be performed as,
\begin{equation}
    \mathbf{I}=\mathit{G}(\beta\cdot\mathbf{z}_1+(1-\beta)\cdot\mathbf{z}_2); {\color{red}\beta\cdot\bm{\theta}_{\mathit{G}_1}+(1-\beta)\cdot\bm{\theta}_{\mathit{G}_2}}),
\end{equation}
where $\bm{\theta}_{\mathit{G}_1}$ and $\bm{\theta}_{\mathit{G}_2}$ are the specialized parameters for reconstructing $\mathbf{x}_1$ and $\mathbf{x}_2$, respectively, and the difference from \cref{eqn:interpolation} is highlighted with {\color{red}red}.

\subsubsection{Style and Attribute Transfer, Image Crossover}
\label{subsubsec:app_style_transfer}
As shown in \cite{StyleGAN}, the features in earlier layers control coarser-grained attributes like structure and pose, while the ones in latter layers control finer-grained attributes like appearance and textures.
For the StyleGAN series, the latent codes are corresponding to these layers in the $\mathcal{W}$ space, and with the style mixing regularization during training, the latent codes in different layers can be distinct~\cite{StyleGAN}.
Therefore, the pre-trained StyleGAN models are ready-to-use for style transfer tasks by replacing the latent codes in latter layers with the ones from the style image~\cite{Image2StyleGAN,StyleGAN2Distillation,guan2020collaborative,GH-Feat,baseline}.
In this way, the characters regarded as style (\eg, color, stroke, texture, \etc) are transferred to the content image.
Instead of simply mixing style codes from different images, some works also perform a more precise control by specifying the attributes to be transferred~\cite{StyleRig,IDDisentanglement}, which can be combined with the methods introduced in \cref{sec:disentanglement} for disentangling the attributes.
Some methods~\cite{pSp,baseline} also considered image-to-image translation tasks such as transferring sketch or parsing (segmentation) images into natural ones, or generating toonified versions of a human face.

\begin{figure}[t]
    \centering
    \scriptsize
    \begin{tabular}{cc}
        \begin{tabular}{ccccc}
            \hspace{-3mm} & \hspace{-3mm} \includegraphics[width=.1\linewidth]{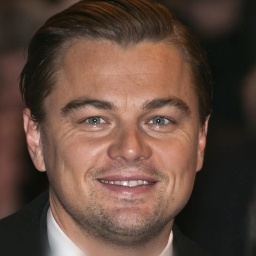} \hspace{-3mm} & \hspace{-3mm} \includegraphics[width=.1\linewidth]{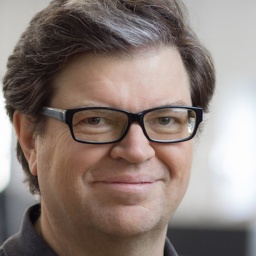} \hspace{-3mm} & \hspace{-3mm} \includegraphics[width=.1\linewidth]{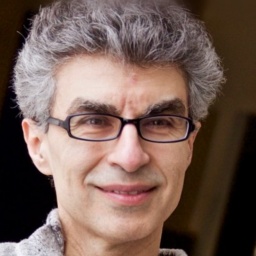} \hspace{-3mm} & \hspace{-3mm} \includegraphics[width=.1\linewidth]{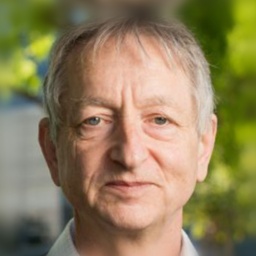} \hspace{-3mm} \\
            \hspace{-3mm} \includegraphics[width=.1\linewidth]{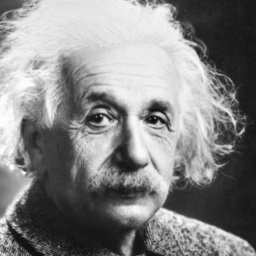} \hspace{-3mm} & \hspace{-3mm} \includegraphics[width=.1\linewidth]{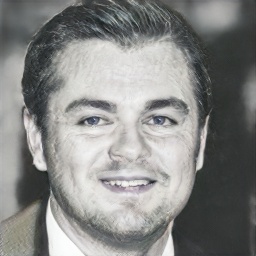} \hspace{-3mm} & \hspace{-3mm} \includegraphics[width=.1\linewidth]{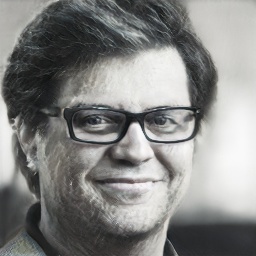} \hspace{-3mm} & \hspace{-3mm} \includegraphics[width=.1\linewidth]{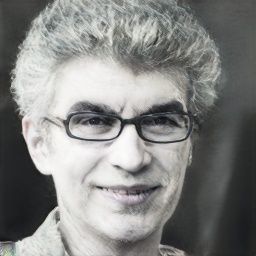} \hspace{-3mm} & \hspace{-3mm} \includegraphics[width=.1\linewidth]{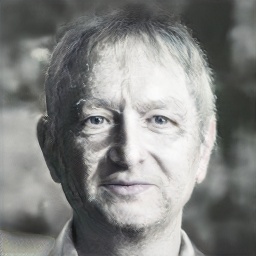} \hspace{-3mm} \\
            \hspace{-3mm} \includegraphics[width=.1\linewidth]{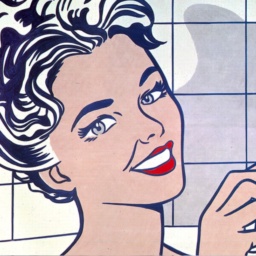} \hspace{-3mm} & \hspace{-3mm} \includegraphics[width=.1\linewidth]{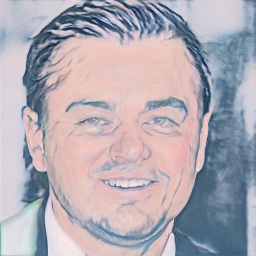} \hspace{-3mm} & \hspace{-3mm} \includegraphics[width=.1\linewidth]{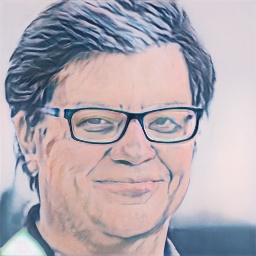} \hspace{-3mm} & \hspace{-3mm} \includegraphics[width=.1\linewidth]{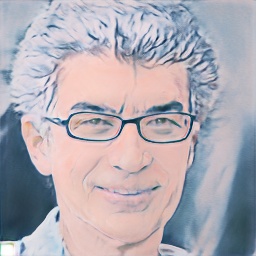} \hspace{-3mm} & \hspace{-3mm} \includegraphics[width=.1\linewidth]{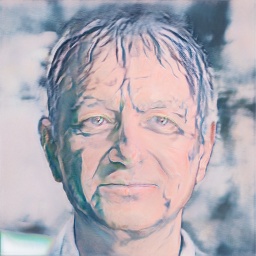} \hspace{-3mm} \\
            \hspace{-3mm} \includegraphics[width=.1\linewidth]{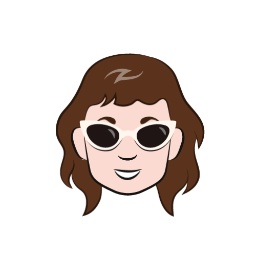} \hspace{-3mm} & \hspace{-3mm} \includegraphics[width=.1\linewidth]{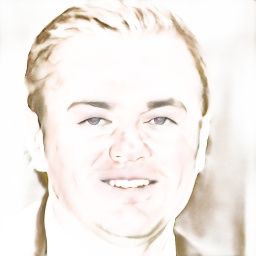} \hspace{-3mm} & \hspace{-3mm} \includegraphics[width=.1\linewidth]{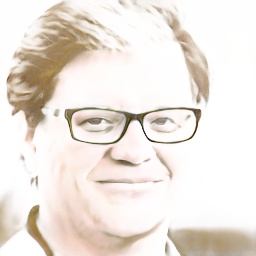} \hspace{-3mm} & \hspace{-3mm} \includegraphics[width=.1\linewidth]{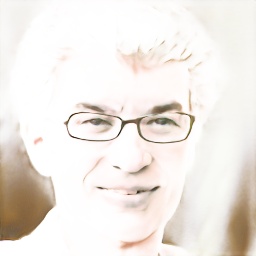} \hspace{-3mm} & \hspace{-3mm} \includegraphics[width=.1\linewidth]{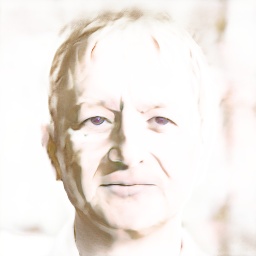} \hspace{-3mm} \\
        \end{tabular} &
        \begin{tabular}{ccc}
            \hspace{-3mm} \includegraphics[width=.1\linewidth]{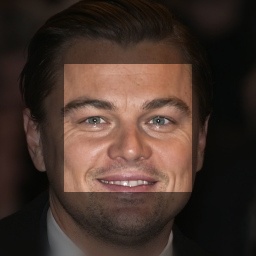} \hspace{-3mm} & \hspace{-3mm} \includegraphics[width=.1\linewidth]{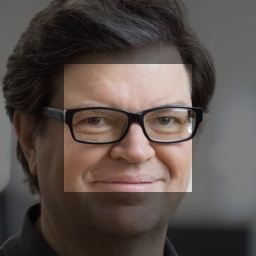}  \hspace{-3mm} & \hspace{-3mm} \includegraphics[width=.1\linewidth]{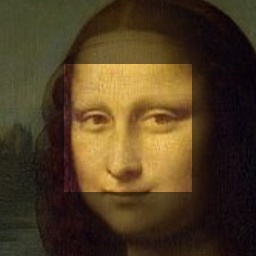} \hspace{-3mm} \\
            \hspace{-3mm} \includegraphics[width=.1\linewidth]{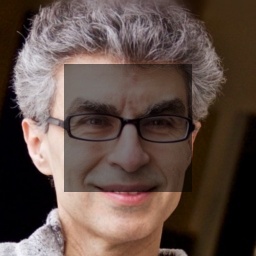}  \hspace{-3mm} & \hspace{-3mm} \includegraphics[width=.1\linewidth]{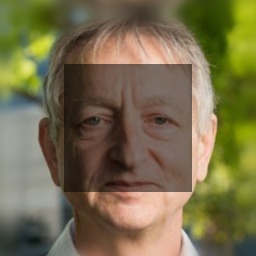}  \hspace{-3mm} & \hspace{-3mm} \includegraphics[width=.1\linewidth]{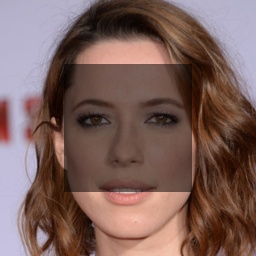} \hspace{-3mm} \\
            \hspace{-3mm} \includegraphics[width=.1\linewidth]{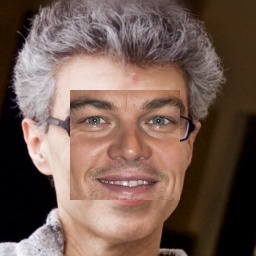}  \hspace{-3mm} & \hspace{-3mm} \includegraphics[width=.1\linewidth]{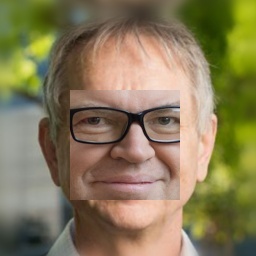}  \hspace{-3mm} & \hspace{-3mm} \includegraphics[width=.1\linewidth]{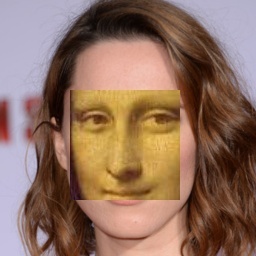} \hspace{-3mm} \\
            \hspace{-3mm} \includegraphics[width=.1\linewidth]{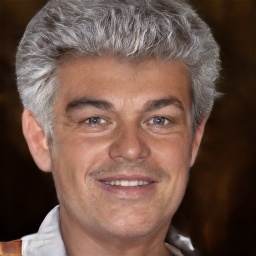}  \hspace{-3mm} &  \hspace{-3mm} \includegraphics[width=.1\linewidth]{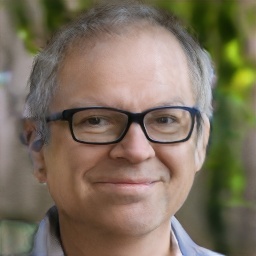}  \hspace{-3mm} & \hspace{-3mm} \includegraphics[width=.1\linewidth]{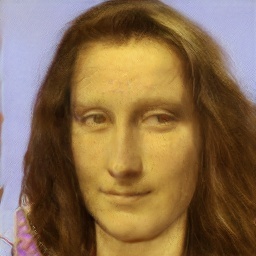} \hspace{-3mm} \\
        \end{tabular} \\[-1mm]
        \hspace{-3mm}Style Transfer & Image Crossover\hspace{-3mm}
    \end{tabular}
    \caption{Style transfer and image crossover. The images are processed by IDInvert~\cite{IDInvert}.}
    \label{fig:app_style_transfer}
\end{figure}

Apart from the global tasks of style and attribute transfer, some methods focus on local modifications from two perspectives, \ie, local style transfer and image crossover.
The former is the style transfer task in a local region, while the latter is to combine multiple images and regenerate the stitched image to make it visually plausible.
For example, \cite{LatentComposition} introduced random masks (\eg, $\mathbf{m}_1$ and $\mathbf{m}_2$) during training, such that the encoder will be compatible with masked input images, making it possible to combine multiple incomplete images by corresponding mask directly,
\begin{equation}
    \mathbf{I}=\mathit{G}(\mathit{E}(\mathbf{m}_1\circ\mathbf{x}_1+\mathbf{m}_2\circ\mathbf{x}_2)),
\end{equation}
which generates more realistic and natural images than blending in pixel or latent spaces like,
\begin{align}
    \mathbf{I}&=\mathit{G}(\mathit{E}(\alpha\cdot\mathbf{x}_1+(1-\alpha)\cdot\mathbf{x}_2)),&(\mathtt{pixel\ space\ blend})\label{eqn:pixel_blend}\\    \mathbf{I}&=\mathit{G}(\alpha\cdot\mathit{E}(\mathbf{x}_1)+(1-\alpha)\cdot\mathit{E}(\mathbf{x}_2)).&(\mathtt{latent\ space\ blend})\label{eqn:latent_blend}
\end{align}
Note that Image2StyleGAN++~\cite{Image2StyleGAN++} performs local style transfer in both $\mathcal{W}+$ and $\mathcal{N}$ spaces. Some methods directly transfer the features rather than latent codes~\cite{suzuki2018spatially,EditingInStyle}.
Visual results can be found in \cref{fig:app_style_transfer}.

\subsubsection{Attribute Editing}
Attribute editing aims to modify certain attributes of an image (\eg, age, gender, \etc), which is a popular application of GAN models in recent years.
The difference between attribute editing and attribute transfer in \cref{subsubsec:app_style_transfer} is that the reference image is not necessary for attribute editing tasks.
Some methods focus on \textit{physical attributes}.
For example, Zhu~\etal~\cite{zhu2016generative} and Abdal~\etal~\cite{Image2StyleGAN++} took color strokes provided by users as input and optimize the image towards a reasonable appearance where the color or shape of regions covered by the strokes are modified, while the background regions are kept unchanged.
Jahanian~\etal~\cite{Steerability} focused on self-supervised training for zooming, shifting, rotating, brightness adjustment, and \etc.

To manipulate the attributes more closely related to \textit{semantic concepts}, some methods~\cite{IcGAN,DGP,GANalyze,StyleGAN2Distillation,StyleFlow,SAM} leveraged pre-trained attribute classifiers or attribute labels of the datasets to identify the manipulation direction in the latent space.
Some methods~\cite{InterFaceGAN,mGANPrior,IDInvert,StyleSpace} chose to first define an attribute separation hyperplane and take the normal vector as editing directions. Considering the non-linearity of the latent space, especially when manipulating the attributes to a large extent, Wang~\etal~\cite{Hijack-GAN} proposed to learn a proxy model to estimate the gradient of the pre-trained GAN models and modify the attributes step by step.
The primary drawback of the aforementioned methods is that they need annotated labels (or classification models trained with these labels) for training.
To eliminate the dependence on these labels, some works~\cite{SeFa,NaviGAN} proposed unsupervised methods to disentangle different attributes, which have been introduced in detail in \cref{sec:disentanglement}.
Besides, some methods~\cite{GANDissection,GANPaint,StyleIntervention} achieved local attribute editing based on the observations introduced in \cref{subsubsec:GANDissection}.

\begin{figure}[t]
    \centering
    \scriptsize
    \begin{tabular}{cc}
        \includegraphics[width=.45\linewidth]{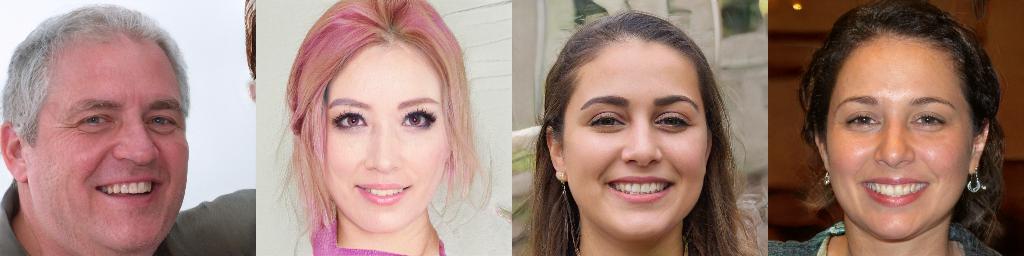} & \includegraphics[width=.45\linewidth]{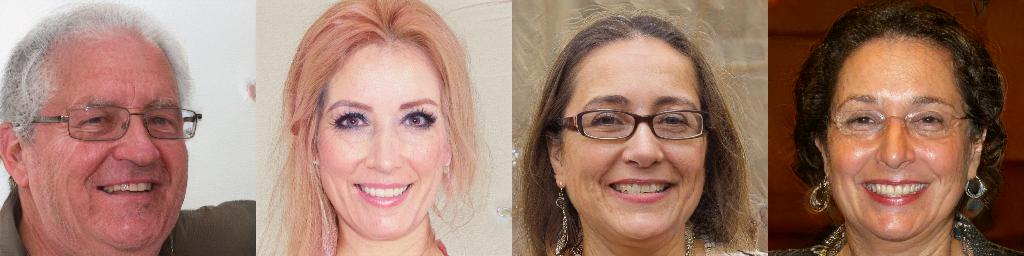} \\[-1mm]
        Input & To \textit{Old} \\
        \includegraphics[width=.45\linewidth]{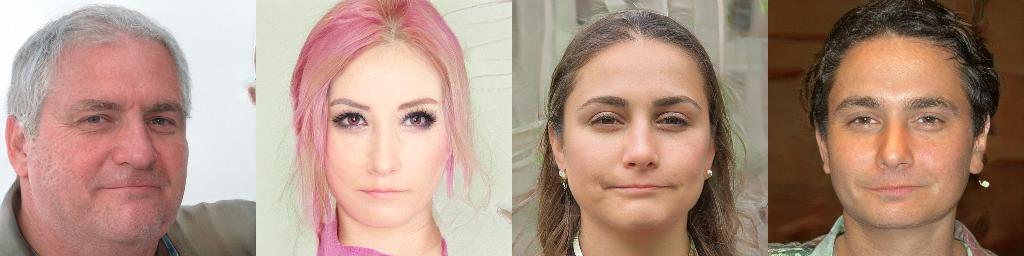} & \includegraphics[width=.45\linewidth]{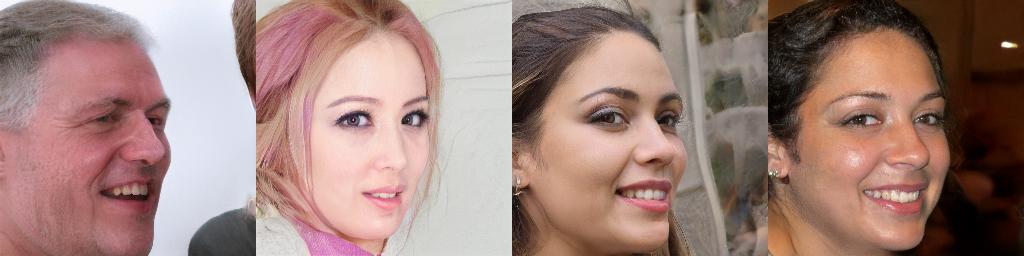} \\[-1mm]
        To \textit{Not Smile} & To \textit{Profile}
    \end{tabular}
    \caption{Attribute editing with manipulation directions found by InterFaceGAN~\cite{InterFaceGAN}.}
    \label{fig:app_attribute_editing}
\end{figure}

There are also some other interesting methods to achieve attribute editing.
Image2StyleGAN~\cite{Image2StyleGAN} uses a one-shot method that obtains the manipulation direction vector via the difference between two images.
\cite{GH-Feat} sampled global or local features in the latent space which results in random new attributes.
E4E~\cite{E4E} proposes to leverage an encoder for generating target latent codes directly instead of the previous inversion+editing methods.
The visual results are given in \cref{fig:app_attribute_editing}.

\subsubsection{Image Transform and Jittering, Visual Alignment}
\label{subsubsec:app_transform}
Since StyleGAN~\cite{StyleGAN} introduces layer-wise noises (\ie, the $\mathcal{N}$ space), it naturally supports random jittering on different layers.
Pan~\etal~\cite{DGP} further added Gaussian noise to the $\mathcal{Z}$ space of BigGAN model~\cite{BigGAN} and achieved image jittering with larger variance.
Jahanian~\etal~\cite{Steerability} also explored the image transformation with pre-trained GANs.
As introduced in \cref{sec:GANs_development}, StyleGAN3~\cite{StyleGAN3} adopts equivariant layers, which greatly facilitates the output image quality when performing transform operations like shifting.

\begin{figure}[t]
    \centering
    \scriptsize
    \begin{tabular}{ccc}
        \hspace{-3mm} \includegraphics[width=.33\linewidth]{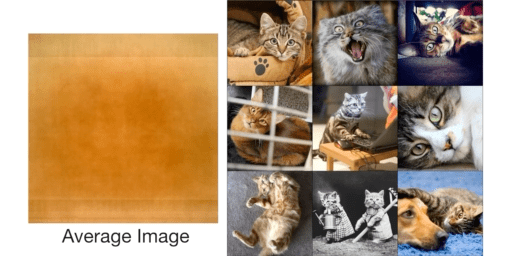} \hspace{-3mm} & \hspace{-3mm} \includegraphics[width=.33\linewidth]{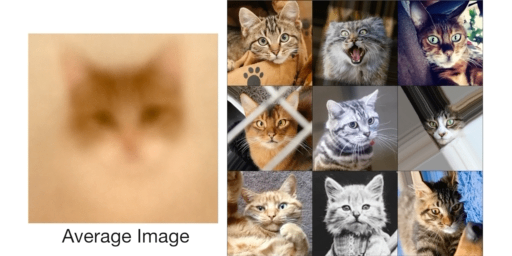} \hspace{-3mm} & \hspace{-3mm} \includegraphics[width=.33\linewidth]{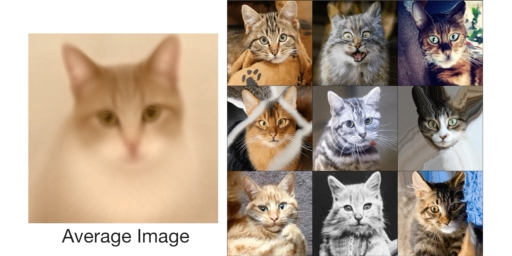}  \hspace{-3mm} \\[-1mm]
        \hspace{-3mm} Averaging Original Images \hspace{-3mm} & \hspace{-3mm} Affine Transform \hspace{-3mm} & \hspace{-3mm} Elastic Transform \hspace{-3mm} \\
        \multicolumn{3}{c}{\includegraphics[width=.95\linewidth]{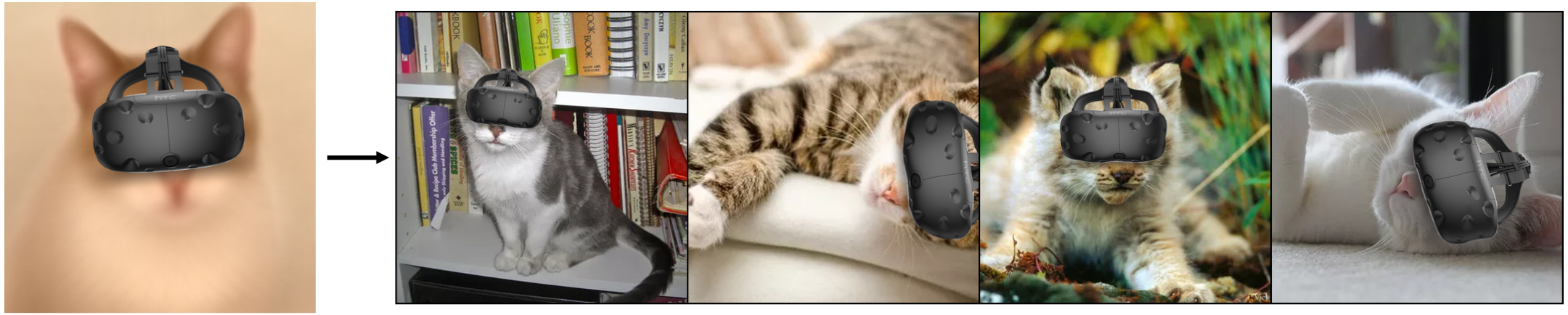}}\\[-1mm]
        \multicolumn{3}{c}{Editing on Aligned Images}
    \end{tabular}
    \caption{Dense visual alignment method~\cite{GANGealing} trained with GAN supervision. Please zoom in for better observation.}
    \label{fig:app_image_transform}
\end{figure}

Based on the image transform operations, some methods leverage pre-trained GANs for visual alignment.
For example, pSp~\cite{pSp} constrains that face images are inverted to the same latent code with its horizontally flipped counterpart, which leads to a frontal face for profile ones.
Peebles~\etal~\cite{GANGealing} further extended the method to more general cases by clustering one or several center points (\eg, frontal for faces), and finally they can obtain a spatial transformer network that can perform visual alignment as well as the reverse process.
In other words, we can edit the transformed (aligned) images for dealing with a certain object in all video frames.
In \cref{fig:app_image_transform}, we show several examples of these applications.

\subsection{Image Restoration}
Unlike the applications introduced in \cref{subsec:app_image_editing}, which mainly leverage the exhilarating latent space disentanglement ability of recent GAN models, image restoration tasks rely more on the high-quality natural image priors provided by the pre-trained large-scale GAN models.
Note that the GAN model is trained to learn the mapping $\mathit{g}:\mathbf{z}\mapsto\mathbf{x}$, where $\mathbf{z}$ is generally sampled from a simple distribution (\eg, the standard Gaussian distribution $\mathbf{z}\sim\mathcal{N}(\mathbf{0}, \mathbf{I})$), while $\mathbf{x}\in\mathcal{X}$ represents the high-quality training datasets.
With the millions or billions of training iterations, we assume that $\{\mathbf{z}_\mathit{i}\}$ covers at least a truncated Gaussian space, or more precisely, a manifold, which is denoted by $\mathcal{M}$.
In other words, given a latent code sampled in $\mathcal{M}$, for a well-trained GAN model helps to constrain the output to the $\mathcal{X}$ space, \ie, the output tends to be a high-quality image following the training image distribution.
In this way, given a corrupted image, we can find a latent code in the latent space, which has the smallest ``distance'' from this corrupted image.
Then the image generated from the latent code can be regarded as the one with the highest likelihood among all possible high-quality counterparts.
The relevant image restoration applications include image super-resolution, image denoising, image inpainting, image colorization, artifacts removal, \etc., and we will go through these applications in this subsection.

\subsubsection{Image Super-resolution, Image Denoising, Image Inpainting, and Image Colorization}
\begin{figure}[t]
    \centering
    \scriptsize
    \begin{tabular}{ccc}
        \hspace{-2.75mm} \includegraphics[width=.32\linewidth]{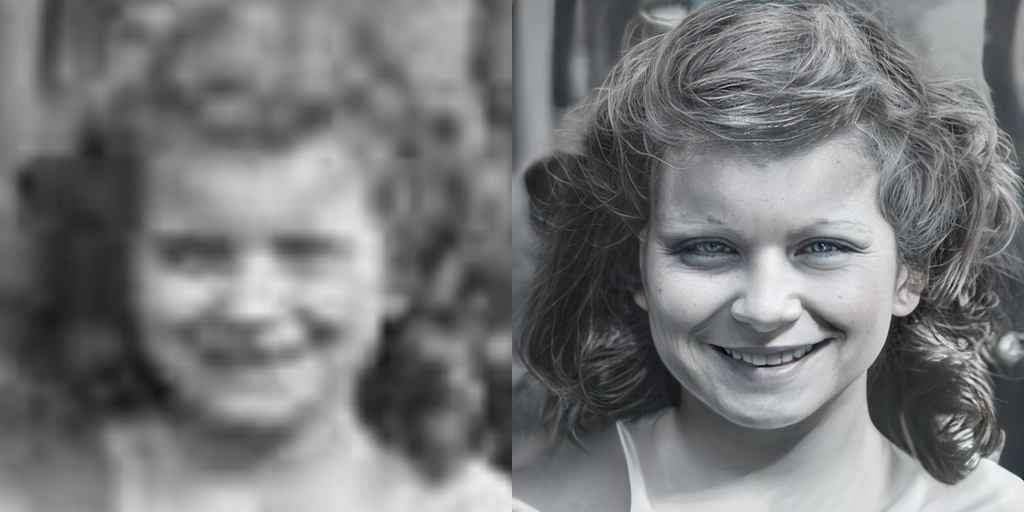} \hspace{-2.75mm} & \hspace{-2.75mm} \includegraphics[width=.32\linewidth]{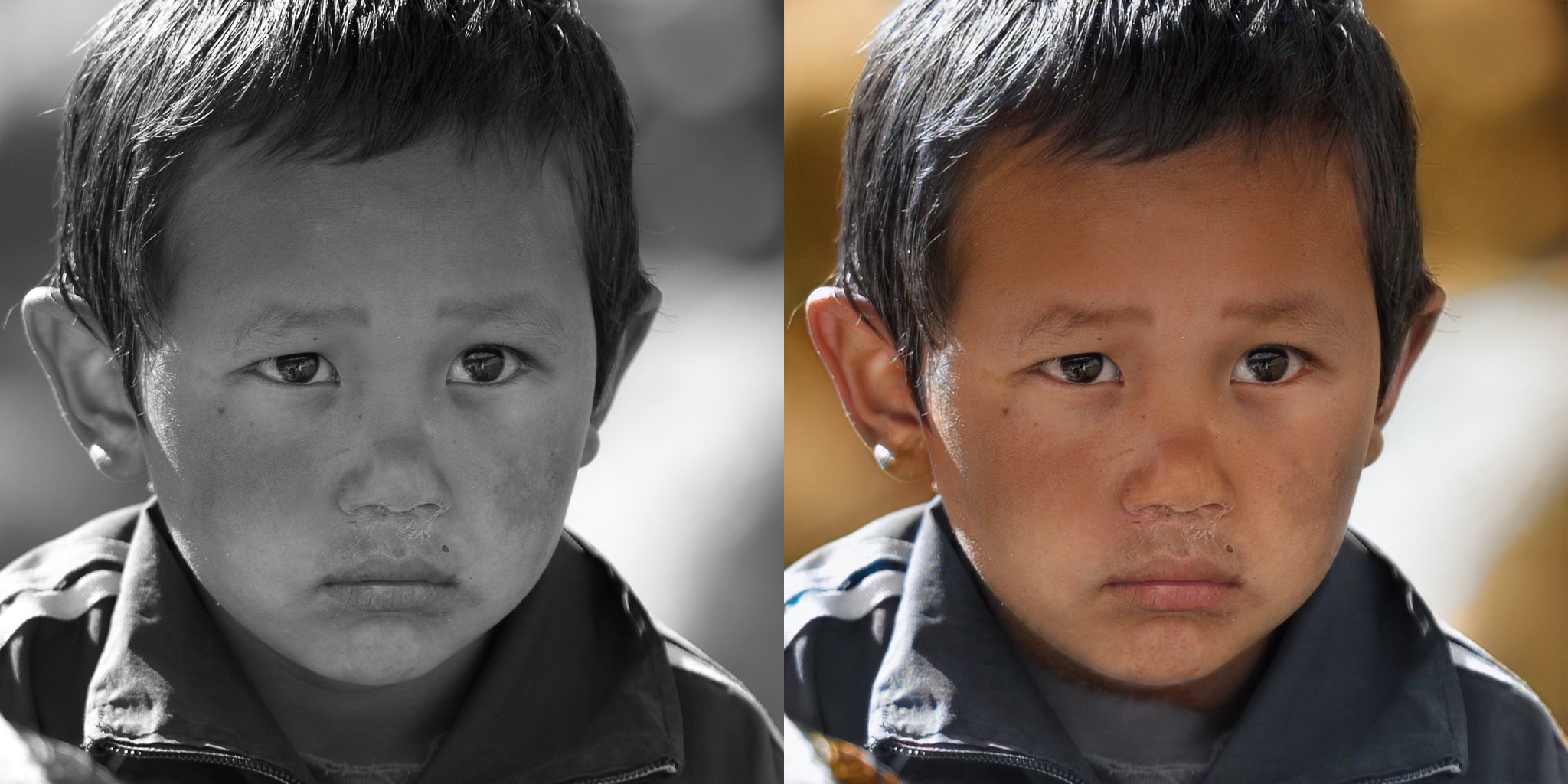} \hspace{-2.75mm} & \hspace{-2.75mm} \includegraphics[width=.32\linewidth]{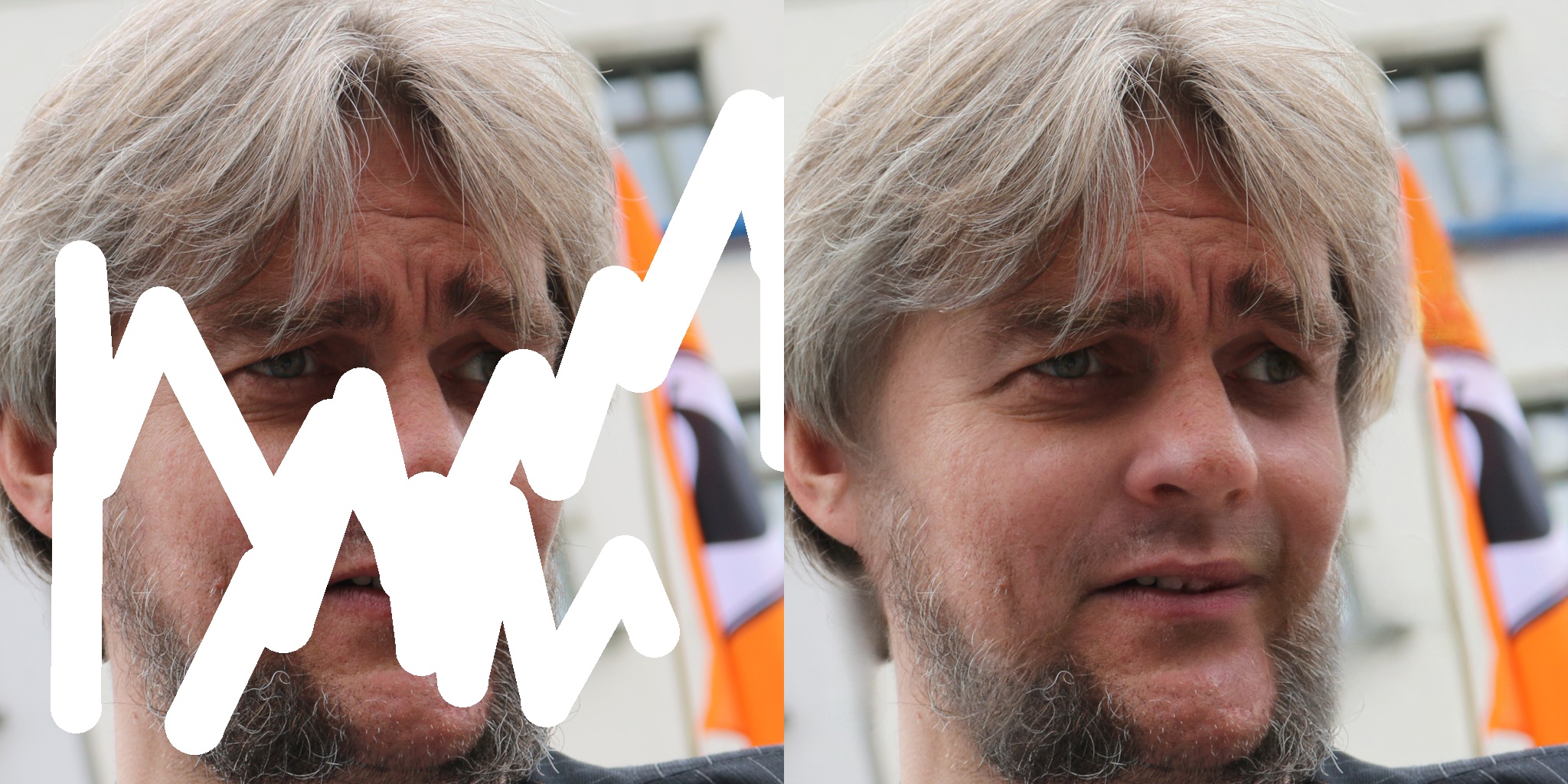} \hspace{-2.75mm}\\[-1mm]
        \hspace{-2.75mm} Real Image Super-resolution \hspace{-2.75mm} & \hspace{-2.75mm} Image Colorization \hspace{-2.75mm} & \hspace{-2.75mm} Image Inpainting \hspace{-2.75mm}
    \end{tabular}
    \caption{Image restoration with GPEN~\cite{GPEN} pre-trained with FFHQ~\cite{StyleGAN} dataset.}
    \label{fig:app_restoration}
\end{figure}

Considering the common methods used for image super-resolution, image denoising, image inpainting, and image colorization, we introduce these applications together in this part.
Overall, these methods can be divided into two classes, \ie, unsupervised and supervised.
Specifically, for unsupervised methods~\cite{mGANPrior,PULSE,DGP}, the optimization objective is typically defined by,
\begin{equation}
    \mathcal{L}_\mathit{unsup}=\argmin_\mathbf{z}\|\zeta(\mathit{G}(\mathbf{z};\bm{\theta}_\mathit{G})) - \mathbf{x}_\mathit{LQ}\|_\mathit{p}^\mathit{p}+\rho(\mathbf{z}),
\end{equation}
where $\zeta$ denotes the degradation function (\eg, down-sampling for image super-resolution, color-to-gray for image colorization, masking for image inpainting), and $\rho$ denotes the regularization on $\mathbf{z}$ derived from the distribution of the latent space (\eg, Gaussian).
Here we note that, as shown in \cref{fig:app_restoration}, the denoising task here is actually inpainting tasks with random defective pixels rather than additive noise suppressing, and for super-resolution tasks the degradation process is approximated by simple down-sampling algorithms like bicubic.
These drawbacks make the optimization-based unsupervised methods limited to a certain range of tasks, where the ground-truth images are unavailable during inference.

To solve this problem, with the development of learning-based methods, the ground-truth (reference) images can be utilized for training the encoder to obtain the restoration results, \ie, the supervised methods.
With the reference, the encoder can directly learn the mapping $\mathit{f}: \mathbf{x}_\mathit{LQ}\mapsto\mathbf{z}_\mathit{HQ}$, which is then delivered into the pre-trained GAN model to obtain the high-quality output image $\mathbf{x}_\mathit{HQ}=\mathit{G}(\mathbf{z}_\mathit{HQ}; \bm{\theta}_\mathit{G})$~\cite{pSp,bartz2020one,GLEAN,ImprovedGANEmbedding,GFPGAN,baseline,GPEN}.
In particular, the loss function is defined by,
\begin{equation}
    \mathcal{L}_\mathit{sup}=\sum_\mathit{i}\|\xi_\mathit{i}(\mathit{G}(\mathit{E}(\mathbf{x}_\mathit{LQ};\bm{\theta}_\mathit{E});\bm{\theta}_\mathit{G}) - \xi_\mathit{i}(\mathbf{x}_\mathit{HQ})\|_\mathit{p}^\mathit{p}+\rho(\mathbf{z}),
\end{equation}
where $\xi_\mathit{i}$ denotes the identity operation or models for loss functions like identity loss~\cite{ArcFace}, LPIPS loss~\cite{LPIPS}, and face parsing loss~\cite{MaskGAN}, which are utilized for better identity preserving or perceptual image quality.
It is worth noting that, unlike other methods which produce the output images directly by the pre-trained GAN model, GLEAN~\cite{GLEAN} stacks an extra decoder on the top of the GAN model, where the features from the pre-trained GAN model as taken as intermediate features to improve the quality of the output generated by the extra decoder.
Visual results can be found in \cref{fig:app_restoration}.

\subsubsection{Artifact Removal}
\begin{figure}[t]
    \centering
    \scriptsize
    \begin{tabular}{cccccc}
        \hspace{-2.75mm} \includegraphics[width=.16\linewidth]{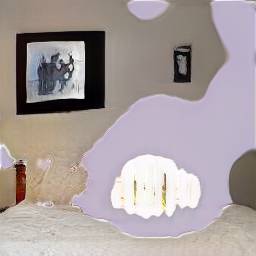} \hspace{-2.75mm} & \hspace{-2.75mm} \includegraphics[width=.16\linewidth]{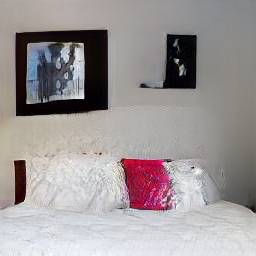} \hspace{-2.75mm} & \hspace{-2.75mm} \includegraphics[width=.16\linewidth]{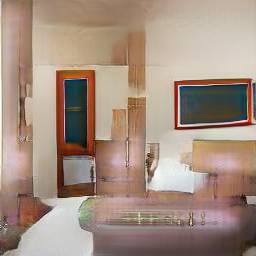} \hspace{-2.75mm} & \hspace{-2.75mm} \includegraphics[width=.16\linewidth]{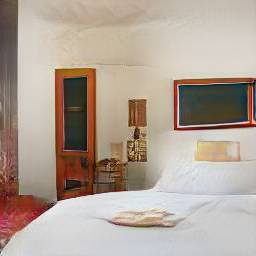} \hspace{-2.75mm} & \hspace{-2.75mm} \includegraphics[width=.16\linewidth]{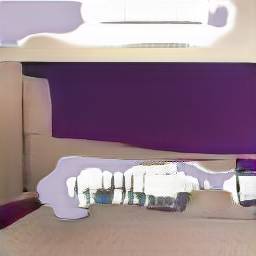} \hspace{-2.75mm} & \hspace{-2.75mm} \includegraphics[width=.16\linewidth]{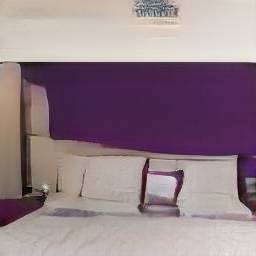}        \hspace{-2.75mm} \\[-1mm]
        \hspace{-2.75mm} Input \hspace{-2.75mm} & \hspace{-2.75mm} Output \hspace{-2.75mm} & \hspace{-2.75mm} Input \hspace{-2.75mm} & \hspace{-2.75mm} Output \hspace{-2.75mm} & \hspace{-2.75mm} Input \hspace{-2.75mm} & \hspace{-2.75mm} Output \hspace{-2.75mm}
    \end{tabular}
    \caption{Artifacts removal via GANDissection~\cite{GANDissection}, where the neurons controlling the generation of artifacts are set to zero.}
    \label{fig:app_artifacts_removal}
\end{figure}

As shown by Bau~\etal~\cite{GANDissection}, similar to other objects, the generation of some artifacts is also controlled by the neurons of the GAN model.
Therefore, the network dissection technique can also be used to identify the neurons that cause the artifacts, which can bring an obvious image quality improvement (the FID~\cite{FID} improves from $\sim$43 to $\sim$27).
Shen~\etal~\cite{InterFaceGAN} further explored artifact removal in the latent space, where a linear-SVM is trained with 4K bad results to obtain the separation hyperplane, and the artifacts are successfully suppressed by using the normal vector.
\cite{GANCorrection} trained an artifact classifier with a proposed dataset, and identified the flawed regions via Grad-CAM~\cite{Grad-CAM}. Then the artifacts are suppressed in a sequential process.
The visual results are shown in \cref{fig:app_artifacts_removal}.

\subsection{Other Applications}
Apart from the aforementioned image editing and restoration applications, the pre-trained GAN models can also be applied to other tasks.
For example, information hiding~\cite{baseline}, unsupervised domain adaptation~\cite{MimicGAN}, adversarial defense~\cite{MimicGAN,DGP}, anomaly detection~\cite{MimicGAN}, 3D reconstruction~\cite{pan20202d,zhang2021unsupervised}, and so on.
Besides, a new trend in leveraging pre-trained GAN models is to combine multiple GAN models for generating different parts (\eg, the face region and the whole body)~\cite{InsetGAN}, which has great potential for generating images with complex scenes and rich contents.

\section{Discussion}
\label{sec:discussion}

\subsection{Challenges and Open Problems}

\subsubsection{GAN Inversion}
For leveraging pre-trained GAN models on real images, it is a fundamental task to invert them into the latent space, with which the images should be well reconstructed via the GAN models.
Although existing GAN inversion methods have demonstrated decent performance, they are still far from perfect, and there are many challenging problems that remain unsolved.

\vspace{0.5em}
\noindent\textbf{Fidelity and Editability}
When inverting a real image into latent space, there exists a trade-off between the reconstruction fidelity and latent code editability.
On the one hand, for methods that only use the $\mathcal{W}$ space (or the derived spaces like $\mathcal{W}+$ or $\mathcal{S}$ spaces), their reconstruction accuracy is greatly limited due to the narrow latent space in comparison to the image space.
Besides, mode dropping~\cite{GANSeeing} is a severe problem influencing the inversion quality.
On the other hand, for methods that leverage spatial dimensions (\eg, the $\mathcal{F}$~\cite{BDInvert}, $\bm{\Theta}$~\cite{NaviGAN,HyperStyle}, and $\mathcal{N}$~\cite{Image2StyleGAN++,GPEN} spaces), the editability becomes a main concern as these spaces are less disentangled.
Wang~\etal~\cite{HFGI} took a step forward by controlling the passage of spatial-dimension features with the latent code in $\mathcal{W}$ space, which achieves a better trade-off between fidelity and editability.
However, it still remains to be better solved for real-world applications, especially when there are geometry changes (\eg, zoom, rotation) or camera movements.

\vspace{0.5em}
\noindent\textbf{Out-of-distribution Generalization}
Since mode dropping~\cite{GANSeeing} occurs even when a subject has shown in the training images, the out-of-distribution generalization ability (on unseen classes or domains) is a main concern of pre-trained GANs.
Indeed, with less constrained optimization, the GAN models are able to generate images from classes that have no intersection with the training set~\cite{Image2StyleGAN}.
Nevertheless, the resultant latent code does not fall into the meaningful latent space, \ie, the semantic directions in the latent space are inconsistent with the out-of-distribution images.
So that we are unable to edit these images following the prior learned in the training set.
Some methods have tried to adapt GAN models to new domains via few-shot learning via cross-domain correspondence~\cite{ojha2021few}, but the properties like editability require further exploration.

\subsubsection{Image Generation, Editing, and Restoration}
Current methods have achieved appealing performance on image generation, editing, and restoration tasks, where the results are photo-realistic and the semantic information is well represented.
Here we conclude some potential directions for better leveraging pre-trained GANs.

\vspace{0.5em}
\noindent\textbf{Multi-GAN Composition}
Due to the limited model capacity and training mechanism, it is rather hard for a single GAN model to generate a complex scene with every object photo-realistic and natural.
Therefore, Fr{\"u}hst{\"u}ck~\etal~\cite{InsetGAN} tried to combine two pre-trained GAN models together for a compositional GAN model, one of which is for generating the whole body of a person, while the other is responsible for providing a better face region.
The two GAN models are combined by joint-optimizing the latent codes.
It is of great potential to generate complex scenes via the combination of multiple GAN models.

\vspace{0.5em}
\noindent\textbf{Physical Rules and 3D Awareness}
For the purpose of image editing or video generation, the GAN models should be aware of the physical rules for realistic results, and the 3D awareness also benefits to produce reasonable structures.
As introduced in \cref{subsubsec:app_transform}, the pre-trained GAN models are able to manipulate the geography properties of the images, in other words, they have implicitly learned some physical rules from the training set.
Besides, the 3D priors (\eg, 3DMM) have been implied for learning face editing models~\cite{StyleRig,PIE}, which shows appealing results.
Furthermore, the embedded physical rules and 3D awareness will also help to combine multiple GAN models for generating complex scenes.

\vspace{0.5em}
\noindent\textbf{Diverse and Reliable Results}
A major problem of existing image editing and restoration methods based on pre-trained GANs is that only one result is obtained for a given input.
Yet these two tasks are both ill-posed, which prefer diverse results for more flexible choices.
More importantly, for image restoration tasks, since the input image is usually greatly degraded, the results are sometimes less reliable.
In this condition, we argue that providing multiple results with confidence scores and/or precise editing directions recommended based on the degraded input will be more applicable, where timely feedback and adjustment can be obtained through user participation (\eg, eyewitness).

\vspace{0.5em}
\noindent\textbf{Multi-modality Combination}
It is common sense that the generated images should be semantically consistent, and we hope that we can directly provide the semantic descriptions to modify the generated (edited, restored) results.
To this end, many works~\cite{StyleClip,DALLE2,imagen} have sought to combine language models with generative models, and exhibit extraordinary performance in text-guided image editing.
It is challenging yet exciting for the pre-trained GAN models to be well-aligned with semantic concepts in the languages, which will make the real-world application based on pre-trained GANs more convenient.

\subsubsection{Others}

\vspace{0.5em}
\noindent\textbf{Combining with Other Generative Models}
Comparing to other generative methods (\eg, VAE~\cite{VAE}, Normalizing Flow~\cite{GLOW}, and Diffusion Models~\cite{DDPM}), the advantages of GAN models lie in the superior image quality (\textit{vs} VAEs), flexible structure (\textit{vs} normalizing flows) and efficient sampling (\textit{vs} diffusion models).
However, GAN models approximate the data distribution in an implicit way, and they are unstable to train and can easily lead to mode collapse.
A potential way is to combine the GAN models with other generative methods, which can give full play to the advantages of both methods.
For example, Lyu~\etal~\cite{lyu2022accelerating} proposed to use GANs to generate an intermediate result for diffusion models, Grover~\etal~\cite{grover2018flow} combined the maximum likelihood of normalizing flow with GANs.

\vspace{0.5em}
\noindent\textbf{Evaluation Metrics}
For boosting the development of GANs and techniques of leveraging pre-trained GAN models, the evaluation metrics play an important role.
For intuitive evaluations such as reconstruction quality and mode dropping, there have been some methods like FID~\cite{FID} and FSD~\cite{GANSeeing}.
However, indirect or more complex evaluations like the detail and semantic consistency after editing and restoration are less explored.
Besides, for the aforementioned challenging problems, corresponding evaluation metrics are also desired to evaluate the effectiveness.

\subsection{Conclusion}
Behind the superior generation ability, GAN models have shown the concept abstraction ability and the learned generative image priors, which have been extensively explored in recent methods.
In this paper, we give a comprehensive survey of recent progress on leveraging pre-trained GAN models for image editing and restoration tasks, where the two main features of recent GAN models are utilized, \ie, the disentanglement ability of the latent space and the generative image priors inherently learned by the GAN models.
By introducing the neuron understanding methods, we try to construct an intuitive and in-depth impression of pre-trained GAN models.
Subsequently, we review the GAN inversion methods as well as the semantic disentanglement and discovery techniques, and the relevant applications of image editing and restoration tasks are introduced.
Finally, we discuss some challenges and open problems in leveraging pre-trained GAN models for image editing and restoration.

\Acknowledgements{This work was supported by National Natural Science Foundation of China (Grant Nos. U19A2073 and 62006064) and Hong Kong RGC RIF (Grant No. R5001-18).}


%
%
%
%
{
    \bibliographystyle{SCIS2022}
    \bibliography{survey}
}

\end{document}